\documentclass{article}

    \PassOptionsToPackage{numbers, compress}{natbib}

\usepackage[preprint]{neurips_2025}

\usepackage{wrapfig}
\usepackage[utf8]{inputenc} 
\usepackage[T1]{fontenc}    
\usepackage{hyperref}       
\usepackage{url}            
\usepackage{booktabs}       
\usepackage{amsfonts}       
\usepackage{nicefrac}       
\usepackage{microtype}      
\usepackage{xcolor}         
\usepackage{graphicx}
\usepackage{multirow}
\usepackage{algorithm}
\usepackage{algpseudocode}  
\usepackage{amsmath}  
\usepackage{enumitem}
\usepackage{wrapfig}

\title{ComfyMind: Toward General-Purpose Generation via Tree-Based Planning and Reactive Feedback}

\author{%
  Litao Guo\thanks{Indicates Equal Contribution} \\
  HKUST(GZ)\\
  \And
  Xinli Xu\footnotemark[1]\\
  HKUST(GZ)\\
  \And
  Luozhou Wang\\
  HKUST(GZ)\\
  \And
  Jiantao Lin\\
  HKUST(GZ)\\
  \And 
  Jinsong Zhou\\
  HKUST(GZ)\\
  \And
  Zixin Zhang\\
  HKUST(GZ)\\
  \And
  Bolan Su\\
  Bytedance\\
  \And
  Ying-Cong Chen\thanks{Indicates Corresponding Author}\\
  HKUST(GZ), HKUST
}

\begin{document}

\maketitle

\begin{figure}[htbp]
  \centering
  \includegraphics[width=1.0\textwidth]{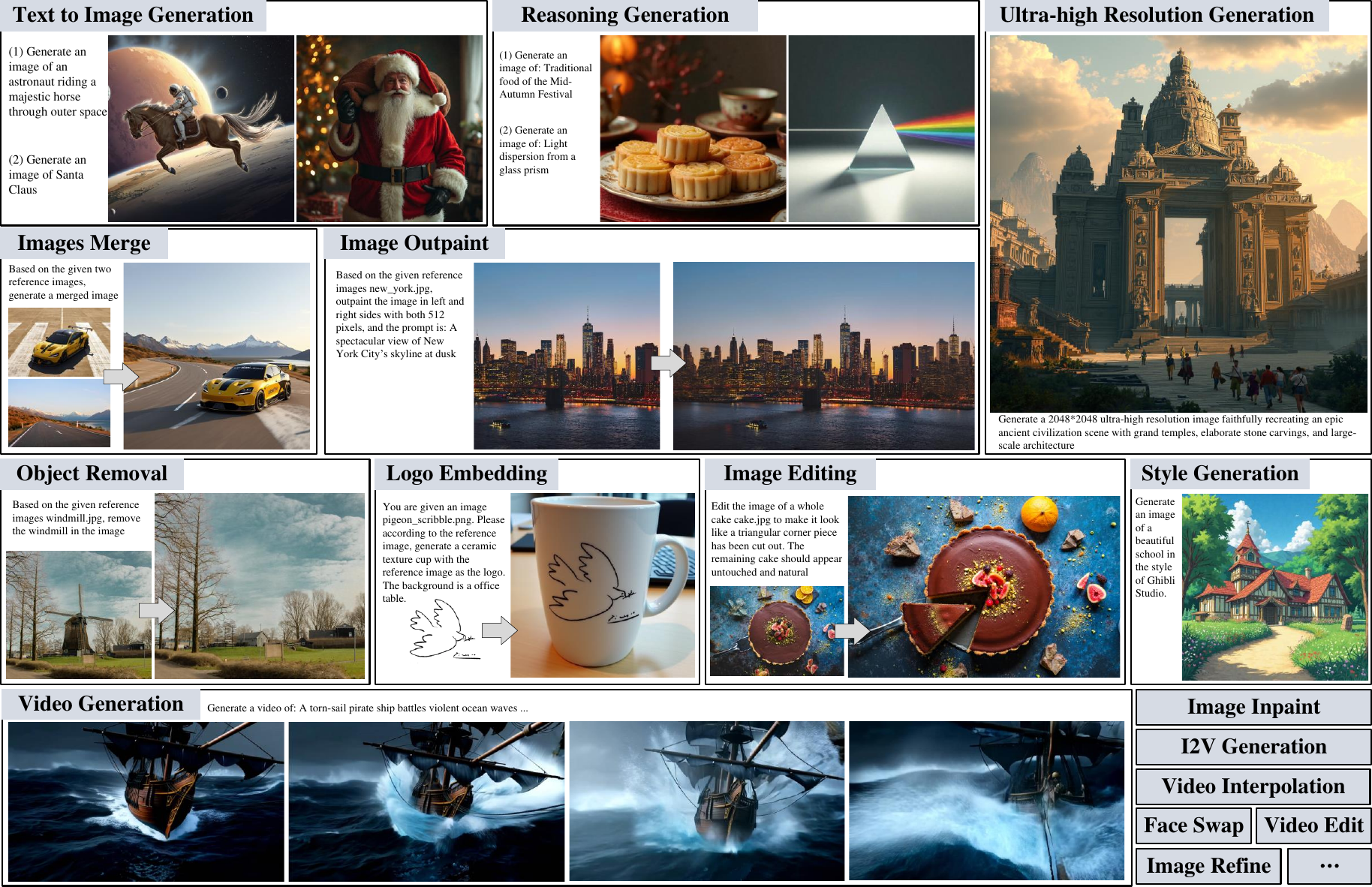}
  \caption{Overview of generative and editing capabilities supported by ComfyMind.}
  \label{fig:topfig}
\end{figure}

\begin{abstract}

With the rapid advancement of generative models, general-purpose generation has gained increasing attention as a promising approach to unify diverse tasks across modalities within a single system. Despite this progress, existing open-source frameworks often remain fragile and struggle to support complex real-world applications due to the lack of structured workflow planning and execution-level feedback. To address these limitations, we present ComfyMind, a collaborative AI system designed to enable robust and scalable general-purpose generation, built on the ComfyUI platform. ComfyMind introduces two core innovations: Semantic Workflow Interface (SWI) that abstracts low-level node graphs into callable functional modules described in natural language, enabling high-level composition and reducing structural errors; Search Tree Planning mechanism with localized feedback execution, which models generation as a hierarchical decision process and allows adaptive correction at each stage. Together, these components improve the stability and flexibility of complex generative workflows. We evaluate ComfyMind on three public benchmarks: ComfyBench, GenEval, and Reason-Edit, which span generation, editing, and reasoning tasks. Results show that ComfyMind consistently outperforms existing open-source baselines and achieves performance comparable to GPT-Image-1. ComfyMind paves a promising path for the development of open-source general-purpose generative AI systems. Project
page:~\url{https://github.com/LitaoGuo/ComfyMind}

\end{abstract}

\section{Introduction} \label{sec:Introduction}

The rapid development of visual generative models has demonstrated remarkable performance across multiple generative tasks, including text-to-image generation~\cite{podell2023sdxl, esser2024scaling, blackforestlabs_flux_2024}, image editing~\cite{brooks2023instructpix2pix, Stability_Ai_Cosxl, huang2024smartedit}, and video generation~\cite{hacohen2024ltx, hunyuanvideo, wang2025wan}. In recent years, research has gradually shifted towards end-to-end general-purpose generative models~\cite{ge2024seed, wu2024janus, fang2025got, gpt_image_1}, aiming to handle these diverse tasks within a single unified model. However, existing open-source general-purpose generative models~\cite{ge2024seed, wu2024janus, fang2025got} still face a series of challenges, including instability in generation quality and a lack of structured planning and composition mechanisms required to handle complex, multi-stage visual workflows. These limitations affect the model's performance and robustness in real-world applications. In contrast, the newly released OpenAI’s GPT-Image-1~\cite{gpt_image_1} has garnered widespread attention for its remarkable capabilities in unified image generations~\cite{yan2025gpt}. However, the closed-source nature of GPT-Image-1 and its primary focus on image generation tasks limit its applicability and scalability across broader multimodal generation tasks.

The ComfyUI platform provides a potential path toward achieving an open-source general-purpose generative approach. ComfyUI is an open-source platform designed to create and execute generative workflows, offering a node-based interface that allows users to construct visual generative workflows represented as JSON according to their needs. The platform's modular design offers high flexibility in constructing workflows. However, despite its flexibility, building complex workflows from scratch remains a challenge, particularly when dealing with customized or intricate task requirements, which demand substantial expertise and considerable time for trial and error. To address this, recent research~\cite{gal2024comfygen,xue2409comfybench} has begun exploring the use of large language models (LLMs) to construct customized workflows, thereby enabling general-purpose visual generation.

\begin{wrapfigure}[14]{r}{0.48\textwidth} 
  \vspace{-1.2em} 
  \centering
  \setlength{\fboxsep}{0pt}
  \setlength{\fboxrule}{0pt}
  \includegraphics[width=\linewidth]{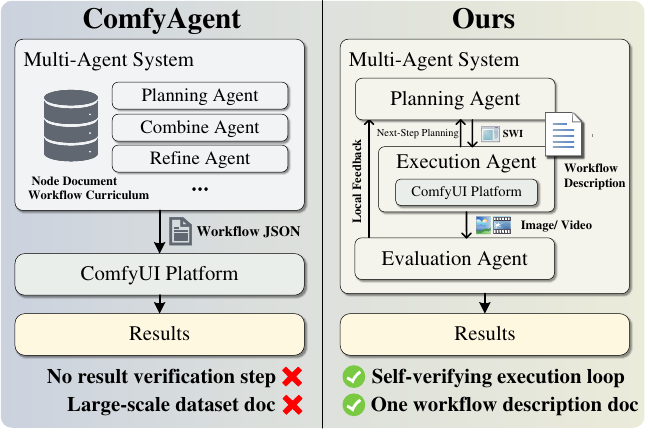}
  \vspace{-1.6em} 
  \caption{\small Structural comparison between ours and ComfyAgent.}
  \label{fig:structural_comp}
\end{wrapfigure}

Building on prior work based on ComfyUI~\cite{comfyui2023}, such as ComfyAgent~\cite{xue2409comfybench}, an automated solution for generating workflows from natural language instructions was proposed. ComfyAgent employs a multi-pronged mechanism to convert natural language instructions into executable workflows. This approach involves pseudo-code translation to convert JSON structures into Python-like code, dynamic retrieval~\cite{lewis2020retrieval}  of node documentation to standardize outputs, and refinement at the workflow level. While this has made significant progress, ComfyAgent also reveals two core issues in low-level workflow generation. First, it treats workflow construction as a flat, token-based decoding task, making it difficult to effectively model modularity and hierarchy. This leads to node omissions, semantic mismatches, and fragile compositions that are challenging to generalize across tasks. Second, the system lacks an execution-level feedback mechanism. Once the workflow is constructed, the system cannot obtain any feedback or error information during generation, hindering incremental corrections and reducing overall robustness.

To address the challenges faced by ComfyAgent, we drew inspiration from human user workflow construction paradigms. We observed that human users typically do not build complex workflows from scratch but instead decompose tasks into smaller subtasks and select appropriate template workflows for each subtask based on higher-level semantics. This modular, step-by-step planning process, combined with localized feedback strategies, allows for incremental refinement and adaptation. When failure occurs, adjustments are made locally rather than globally. This hierarchical planning and reactive feedback strategy enhances the ability to solve complex problems and increases robustness.

In this work, we simulate this human-like strategy and propose a novel framework called \textbf{ComfyMind}. As shown in Figure~\ref{fig:topfig}, our framework exhibits strong generality, supporting a wide range of image and video generation and editing tasks. The framework represents workflow generation as the semantic combination of template workflows, rather than the token-based synthesis of node configurations. Specifically, ComfyMind treats template workflows as atomic semantic modules, each with a clearly defined function, input/output interfaces, and natural language descriptions. By reasoning over these higher-level components, ComfyMind achieves more stable and controllable task compositions. In addition, the semantic-level abstraction also enables effortless integration of new workflows, allowing ComfyMind to quickly adapt to emerging community-contributed models and tasks.


ComfyMind consists of two core mechanisms. The first is the \textbf{Semantic Workflow Interface (SWI)}, which abstracts low-level node graphs into callable semantic functions annotated with structured inputs, outputs, and natural-language captions. This abstraction allows language models to operate on workflows at the semantic level, reducing exposure to platform-specific syntax and minimizing structural errors. The second mechanism is a \textbf{Search Tree Planning with Local Feedback Execution}, which models task execution as a hierarchical decision process. Each node in the planning tree represents a sub-task, and each edge corresponds to a selected SWI module. During execution, failures trigger localized corrections at the current tree layer, avoiding full-chain regeneration and significantly improving robustness. The comparison with previous work is shown in Figure ~\ref{fig:structural_comp}.

We evaluate ComfyMind on three public benchmarks: ComfyBench, GenEval, and Reason-Edit. On ComfyBench, ComfyMind improves the workflow pass rate from 56.0\% to 100\%, and the task resolution rate from 32.5\% to 83.0\%. On GenEval, it achieves an overall performance score of 0.90, outperforming all baseline methods including GPT-Image-1. On Reason-Edit, it reaches a GPT-score of 0.906, surpassing all open-source systems and matching the performance of proprietary models. 

\paragraph{Our contributions are summarized as follows:}
\begin{itemize}
    \item We introduce \textbf{ComfyMind}, a general-purpose generation framework that conceptualizes visual content creation as a modular, semantically structured planning process, and can be instantiated on node-based execution systems (e.g., ComfyUI) that support modular composition and hierarchical task planning.
    \item We propose a unified control mechanism that integrates a Semantic Workflow Interface for high-level modular abstraction with Search Tree Planning with Local Feedback Execution, enabling semantically grounded composition, adaptive correction, and robust realization of complex multi-stage workflows.

    \item We validate ComfyMind on three public benchmarks: ComfyBench, GenEval, and Reason-Edit, covering tasks in generation, editing, and reasoning. The results demonstrate strong performance, broad task generality, and the potential of semantic workflow composition as a foundation for general-purpose visual generation.

\end{itemize}

\section{Related work} \label{sec:Relatedwork}


\subsection{General-Purpose Visual Generation}

Traditional visual generative models typically design task-specific architectures~\cite{zhang2023adding, ye2023ip, gal2022image} to address various generation tasks. With the rapid development of generation models, many studies have shifted towards general-purpose generation models, aiming to solve a wide range of generation tasks using a single model~\cite{ma2024janusflow, team2024chameleon, wang2024emu3, wu2024janus, wu2024vila, xie2024show, zhao2024monoformer, xiao2024omnigen, chen2025janus}. These approaches generally unify image understanding and image generation within one model, enabling more efficient and consistent performance across different tasks. Recently, OpenAI’s closed-source model GPT-Image-1~\cite{gpt_image_1} has gained significant attention due to its remarkable capabilities in image generation, editing, and modification, achieving top performance across multiple domains. 

This study aims to leverage the open-source platform ComfyUI to design a collaborative AI system. By constructing flexible workflows for each task, we strive to achieve general-purpose visual generation. Compared to current general-purpose generation methods, our approach offers distinct advantages: it not only encompasses video modalities but also effectively handles the structured planning and integration of complex, multi-stage visual workflows.

\subsection{Planning and Feedback Mechanisms in Collaborative AI Systems}

In collaborative AI systems, planning and feedback mechanisms are essential for efficiently managing complex tasks. Early systems like AutoGPT~\cite{richards2023autogpt} generate static goal lists with minimal intermediate feedback, often leading to inefficiencies when sub-goals fail. In contrast, more dynamic systems like AutoGen~\cite{wu2023autogen} allow for iterative plan adjustments through dialogue, enhancing adaptability. OpenAgents~\cite{xie2023openagents} provides structured execution pipelines tailored to domain-specific tasks but at the cost of flexibility. ReAct~\cite{yao2023react} introduces a reasoning-acting loop that offers fine-grained control, though it may not be ideal for more complex workflows that require extensive coordination across multiple agents. MetaGPT~\cite{hong2023metagpt} models human-team roles to standardize collaboration, but it lacks the flexibility needed for dynamic, evolving tasks.

Despite these advancements, existing systems still face challenges in localized error handling and recovery. Many systems lack mechanisms for evaluating intermediate outcomes, making them prone to failure in dynamic, non-linear workflows, such as those encountered in generative tasks like ComfyUI~\cite{xue2409comfybench}. Motivated by these limitations, we propose a hierarchical search-tree planning strategy combined with local feedback execution to ensure more robust adaptation and fine-grained error recovery during task execution.

\section{ComfyMind} \label{sec:method}


\subsection{ComfyUI Platform} \label{sec:ComfyUIplatform}

ComfyUI is an open-source platform for designing and executing generative pipelines. User-defined workflows are encoded as JSON-based directed acyclic graphs (DAGs) and interpreted node-by-node on the server to produce images or videos. This node-level abstraction significantly lowers the entry barrier for human artists by modularizing the execution process into visual, interpretable components.

ComfyUI, with the efforts of the open-source community, has integrated support for a wide range of advanced models, covering various modalities from image generation to video production. Notable models include Flux~\cite{blackforestlabs_flux_2024}, Flux Redux~\cite{blackforestlabs_flux_redux_2024}, ACE++~\cite{mao2025ace++}, Segment Anything~\cite{kirillov2023segment, ravi2024sam}, Hunyuan~\cite{hunyuanvideo}, and Wan~\cite{wang2025wan}, among others. This broad support enables users to experiment with state-of-the-art tools for more diverse and sophisticated creative processes. In contrast,  building complex workflows from scratch remains a significant challenge. Constructing workflows that meet highly customized or complex task requirements not only requires deep technical knowledge but also demands significant time and trial-and-error, creating considerable barriers for users trying to efficiently develop new workflows or address specific needs.

\subsection{Semantic Workflow Interface} \label{sec:swi}

ComfyAgent aims to construct demand-customized ComfyUI workflows through LLMs to achieve a general approach to visual generation. It has meticulously built a large-scale Node document dataset, designed an intricate collaborative AI system, employed pseudocode to replace JSON for enhanced workspace representation, and utilized RAG technology. However, it still faces issues such as syntax errors in workflows and missing key Nodes, which result in a pass rate of only 56\% on ComfyBench. The fundamental reason for this low pass rate lies in the vulnerability of low-level representations to LLMs and the conceptualization of workflow construction as a planar, token-based decoding task, which makes it difficult to effectively model modularity and hierarchy.

In contrast to ComfyAgent's paradigm of building entire workflows at the low-level, we adopt a human-like approach to workflow construction by decomposing generation tasks into modular subtasks, each handled independently by a planning agent. Within each subtask, the planning agent selects the most appropriate atomic workflow from the workflow library as a tool. Unlike complex workflows, each atomic workflow is responsible for a simple, single-step generation process, such as text-to-image generation or mask generation. In other words, we replace the single token in ComfyAgent with atomic workflows as the minimal unit in workflow construction.

Following this approach, we introduce a semantic workflow interface, which uses natural language functions instead of low-level JSON specifications as an intermediate representation for workflow construction. Each atomic workflow, encapsulating a function, is annotated with a simple natural language description outlining its purpose, required parameters, and usage. Based on this metadata, the planning agent in ComfyMind selects the most appropriate function for invocation. During the invocation, required parameters (e.g., prompts or reference images) and optional high-level constraints are passed. The execution agent then maps the selected function to its corresponding JSON representation, injecting the parameters. Finally, LLMs perform adaptive parameter-level tuning on the JSON to meet additional constraints. The resulting workflow is executed via the ComfyUI platform, thus completing the generation of the individual subtask.

This abstraction allows the LLM to operate entirely at the semantic level, bypassing the complexities of low-level syntactic grammar and the difficulty of effectively modeling modularity and hierarchy. By eliminating this bottleneck, ComfyMind significantly enhances the robustness of execution. The SWI also minimizes the reliance on fine-grained node documentation. While ComfyAgent’s operation depends on a meticulously crafted dataset containing 3,205 distinct node descriptions, ComfyMind only requires a single unified document to describe the available atomic workflows. Without the need for RAG, ComfyMind can directly inject workflow metadata into the LLM’s context window, ensuring full visibility and eliminating the dependency on external lookups. Ultimately, this reduction in documentation facilitates the seamless integration of newly developed or task-specific workflows. This design enables ComfyMind to quickly integrate emerging workflows from the broader ComfyUI community, while allowing users the flexibility to customize workflow documentation and repositories to meet specific needs.

\begin{figure}[htbp]
  \centering
  \includegraphics[width=1.0\textwidth]{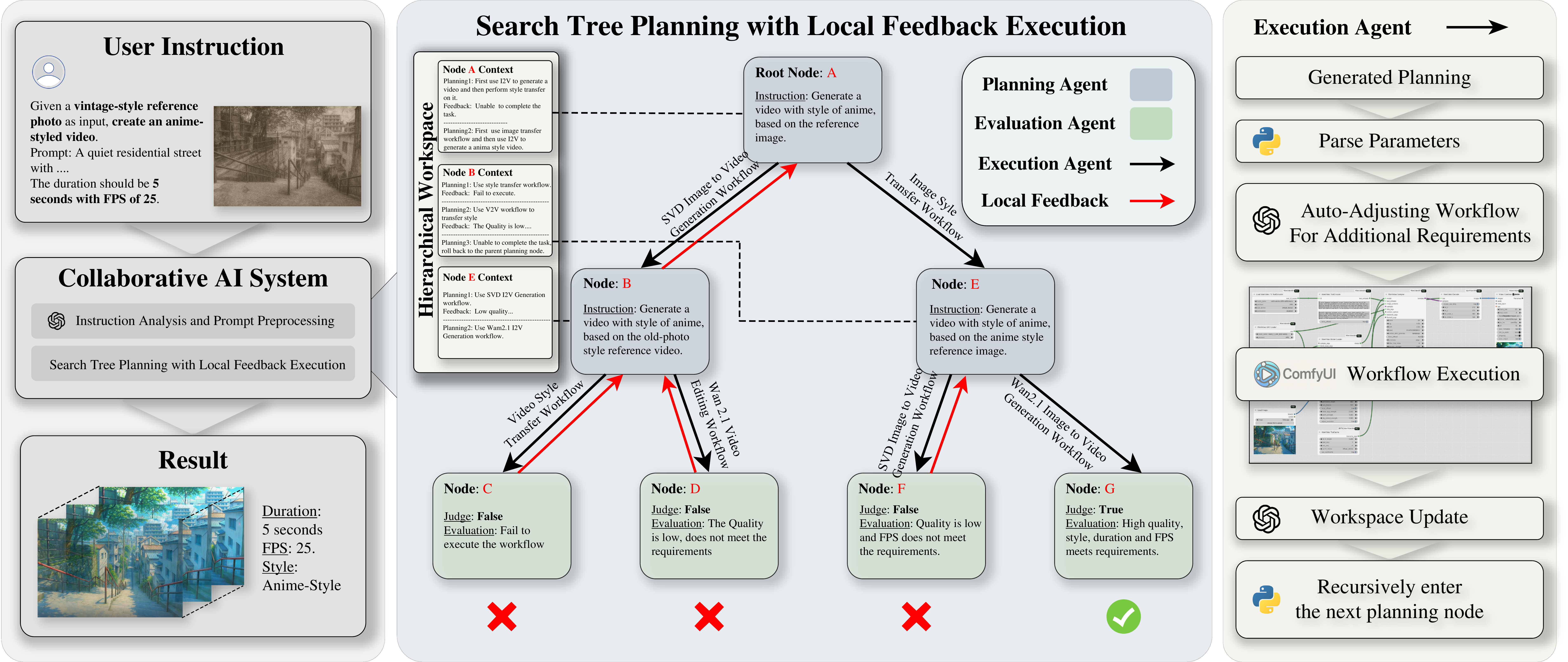}
  \vspace{-1.0em} 
  \caption{Overview of ComfyMind pipeline. Given a user instruction, the system first parses the task and delegates it to Planning Agent. The Agent incrementally explores a semantic search tree, where each node proposes a candidate workflow and receives local feedback based on execution results.}
  \label{fig:pipeline}
\end{figure}

\subsection{Search Tree Planning with Local Feedback Execution} \label{sec:tree}

As introduced in Section~\ref{sec:swi}, SWI enables LLMs to invoke community-validated atomic workflow using natural-language function calls. However, the system must still determine: \textit{how to compose multiple SWI calls into a coherent and task-completing program}. To address this, as sketched in Figure~\ref{fig:pipeline}, ComfyMind introduces a mechanism we term \textbf{Search Tree Planning with Local Feedback Execution}, which formulates workflow construction as a search process over a semantic planning tree. In this structure, each node represents a local planning agent responsible for a specific sub-task, while each edge denotes an execution agent that invokes an SWI function and propagates the result. A complete path from the root to a leaf yields the final visual output that satisfies the user instruction.

At each planning node, the agent examines the current hierarchical workspace state—including text, images, context, and the available workflow documentation. Based on this information, it generates a chain of SWI functions aimed at advancing the current task. Only the first function in the chain is executed at this stage, with its arguments passed to the execution agent. This transition is equivalent to following an edge in the planning tree.

The execution agent translates the selected function into its canonical JSON form as defined by the SWI, applies lightweight parameter adjustments based on higher-level constraints, and executes the workflow using the ComfyUI platform. Throughout the process, the underlying DAG structure is preserved to ensure syntactic correctness. After execution, a Vision-Language Model (VLM) parses and annotates the generated visual content. The resulting artifact, its semantic description, and the updated task specification collectively define the workspace for the next planning node.

If the planning agent determines that its sub-task can be completed with a single operation, it issues a termination signal and invokes the evaluation agent to assess the final output in terms of semantic alignment and perceptual quality. If the result passes evaluation, the search ends successfully. Otherwise, a failure signal and diagnostic feedback are passed to the parent node, which records the outcome and revises its planning strategy accordingly. If no viable options remain at the current level, the error signal propagates upward. Crucially, all feedback is strictly confined to the current hierarchy level, preventing global rollback and preserving valid partial results.

Compared to the step-by-step observe-then-act execution style of ReAct~\cite{yaoreact} planners, our method offers complete history tracking and structured backtracking capabilities. This allows the system to roll back only to the most recent viable decision point upon failure, rather than restarting the entire process—thus avoiding redundant recomputation. At the same time, it improves planning stability by preventing repeated re-planning cycles caused by the lack of stable intermediate states, which can otherwise lead to strategy oscillations and failure to converge.

\section{Experiments} \label{sec:Experiments}

To assess our system’s generative capabilities, we conduct a three‑pronged evaluation. \textbf{ComfyBench}~\cite{xue2409comfybench} quantifies the system’s ability to autonomously build workflows and general-purpose generation; \textbf{GenEval}~\cite{ghosh2023geneval} evaluates the system’s T2I generation capabilities; \textbf{Reason‑Edit}~\cite{huang2024smartedit} measures how well complex editing instructions are executed. Experiments demonstrate that our method surpasses the strongest open‑source baselines by a substantial margin across all three benchmarks and achieves performance comparable to GPT-Image-1. An ablation study further confirms the contribution of each design component. More experimental results are presented in Appendix~\ref{sec:Supplementary_Experiments}.

\subsection{Autonomous Workflow Construction}

\begin{table}[htbp]
  \centering
  \caption{Evaluation of Autonomous Workflow Construction on ComfyBench~\cite{xue2409comfybench}.}
  \label{tab:comfybench}
  \resizebox{\linewidth}{!}{
    \begin{tabular}{l|cccccccc}
      \toprule
      \multirow{2}{*}{\textbf{Agent}} & \multicolumn{2}{c}{\textbf{Vanilla}} & \multicolumn{2}{c}{\textbf{Complex}} & \multicolumn{2}{c}{\textbf{Creative}} & \multicolumn{2}{c}{\textbf{Total}} \\
      \cmidrule(lr){2-3} \cmidrule(lr){4-5} \cmidrule(lr){6-7} \cmidrule(lr){8-9}
      & \%Pass & \%Resolve & \%Pass & \%Resolve & \%Pass & \%Resolve & \%Pass & \%Resolve \\
      \midrule
      GPT-4o + Zero-shot        & 0.0  & 0.0  & 0.0  & 0.0  & 0.0  & 0.0  & 0.0  & 0.0  \\
      GPT-4o + Few-shot~\cite{brown2020language}         & 32.0 & 27.0 & 16.7 & 8.3  & 7.5  & 0.0  & 22.5 & 16.0 \\
      GPT-4o + CoT~\cite{wei2022chain}              & 44.0 & 29.0 & 11.7 & 8.3  & 12.5 & 0.0  & 28.0 & 17.0 \\
      GPT-4o + CoT-SC~\cite{wang2022self}           & 45.0 & 34.0 & 11.7 & 5.0  & 15.0 & 0.0  & 29.0 & 18.5 \\
      Claude-3.5-Sonnet + RAG~\cite{lewis2020retrieval}   & 27.0 & 13.0 & 23.0 & 6.7  & 7.5  & 0.0  & 22.0 & 8.5  \\
      Llama-3.1-70B + RAG       & 58.0 & 32.0 & 23.0 & 10.0 & 15.0 & 5.0  & 39.0 & 20.0 \\
      GPT-4o + RAG              & 62.0 & 41.0 & 45.0 & 21.7 & 40.0 & 7.5  & 52.0 & 23.0 \\
      o1-mini + RAG             & 32.0 & 16.0 & 21.7 & 8.3  & 12.5 & 7.5  & 25.0 & 12.0 \\
      o1-preview + RAG          & 70.0 & 46.0 & 48.3 & 23.3 & 30.0 & 12.5 & 55.5 & 32.5 \\
      \midrule
      ComfyAgent~\cite{xue2409comfybench}       & 67.0 & 46.0 & 48.3 & 21.7 & 40.0 & 15.0 & 56.0 & 32.5 \\
      \textbf{Ours}       & \textbf{100.0} & \textbf{92.0} & \textbf{100.0} & \textbf{85.0} & \textbf{100.0} & \textbf{57.5} & \textbf{100.0} & \textbf{83.0} \\
      \bottomrule
    \end{tabular}
  }
\end{table}

We assessed the autonomous workflow construction capacity of our method in ComfyBench~\cite{xue2409comfybench}. ComfyBench comprises 200 graded difficulty generative and editing tasks that span image and video modalities. For each task, the agent must synthesize workflows that can be executed by ComfyUI. The benchmark reports (i) a \emph{pass rate}, reflecting whether the workflow is runable, and (ii) a \emph{resolve rate}, reflecting whether the output satisfies all task requirements.

As shown in Table~\ref{tab:comfybench}, powered by SWI, our system achieves a \textbf{100\% pass rate} across all difficulty tiers. Our methods eliminate JSON‑level failures that still impede the strongest baseline, ComfyAgent.

More importantly, the proposed Search‑Tree Planning with Local‑Feedback Execution delivers substantial gains in task resolution: relative to ComfyAgent, \textbf{Resolve rate increases by 100\%, 292\%, and 283\% on the Vanilla, Complex, and Creative subsets}, respectively.
Appendix further demonstrates that our system successfully addresses a wide spectrum of user instructions.
The strong generalization ability and output quality evidenced here point to multi-agent systems based on ComfyUI as a promising avenue toward general-purpose generative AI.

\subsection{Text-to-Image Generation}

\subsubsection{Quantitative Results}
We evaluate our system's capability in T2I generation using GenEval~\cite{ghosh2023geneval}. GenEval measures compositional fidelity across six dimensions, including single or two objects, count, color accuracy, spatial positioning, and attribute binding. We compare our method against three strong categories of baselines: (i) \emph{Frozen Text Encoder Mapping Methods}, represented by SD3; (ii) \emph{LLM/MLLM-enhanced methods}, such as Janus and GoT; and (iii) OpenAI’s recently released GPT-Image-1.

As shown in Table~\ref{tab:t2i-geneval}, our system achieves an overall score of \textbf{0.90}, benefiting from its integration of prompt optimization workflows and local feedback execution. This result surpasses all baselines by +0.16 over SD3 and +0.10 over Janus-Pro-7B. Moreover, our system exceeds GPT-Image-1 in five out of six dimensions and the overall score. These results demonstrate that our ComfyUI-based system not only offers strong generality but also is capable of consolidating the strengths of diverse open models, achieving state-of-the-art performance in image synthesis.

\begin{table}[htbp]
  \centering
  \caption{Evaluation of T2I generation on GenEval~\cite{ghosh2023geneval}. Obj.: Object. Attr.: Attribution.}
  \label{tab:t2i-geneval}
  \resizebox{\linewidth}{!}{
      \begin{tabular}{l|c|cccccc} 
        \toprule
        \textbf{Method}  & \textbf{Overall} & \textbf{Single Obj.} & \textbf{Two Obj.} & \textbf{Counting} & \textbf{Colors} & \textbf{Position} & \textbf{Attr. Binding} \\
        \midrule
        \multicolumn{8}{l}{\textit{Frozen Text Encoder Mapping Methods}} \\
        SDv1.5~\cite{rombach2022high}                  & 0.43 & 0.97 & 0.38 & 0.35 & 0.76 & 0.04 & 0.06 \\
        SDv2.1~\cite{rombach2022high}                  & 0.50 & 0.98 & 0.51 & 0.44 & 0.85 & 0.07 & 0.17 \\
        SD-XL~\cite{podell2023sdxl}                    & 0.55 & 0.98 & 0.74 & 0.39 & 0.85 & 0.15 & 0.23 \\
        DALLE-2~\cite{ramesh2022hierarchical}          & 0.52 & 0.94 & 0.66 & 0.49 & 0.77 & 0.10 & 0.19 \\
        SD3-Medium~\cite{esser2024scaling}             & 0.74 & 0.99 & 0.94 & 0.72 & 0.89 & 0.33 & 0.60 \\
        \midrule
        \multicolumn{8}{l}{\textit{LLMs/MLLMs Enhanced Methods}} \\
        LlamaGen~\cite{sun2024autoregressive}                   & 0.32 & 0.71 & 0.34 & 0.21 & 0.58 & 0.07 & 0.04 \\
        Chameleon~\cite{team2024chameleon}                      & 0.39 & -    & -    & -    & -    & -    & -    \\
        LWM~\cite{liu2024world}                                 & 0.47 & 0.93 & 0.41 & 0.46 & 0.79 & 0.09 & 0.15 \\
        SEED-X~\cite{ge2024seed}                                & 0.49 & 0.97 & 0.58 & 0.26 & 0.80 & 0.19 & 0.14 \\
        Emu3-Gen~\cite{wang2024emu3}                            & 0.54 & 0.98 & 0.71 & 0.34 & 0.81 & 0.17 & 0.21 \\
        Janus~\cite{wu2024janus}                                & 0.61 & 0.97 & 0.68 & 0.30 & 0.84 & 0.46 & 0.42 \\
        JanusFlow~\cite{ma2024janusflow}                        & 0.63 & 0.97 & 0.59 & 0.45 & 0.83 & 0.53 & 0.42 \\
        Janus-Pro-7B~\cite{chen2025janus}                        & 0.80 & 0.99 & 0.89 & 0.59 & 0.90 & \textbf{0.79} & 0.66 \\
        GoT~\cite{fang2025got}                                  & 0.64 & 0.99 & 0.69 & 0.67 & 0.85 & 0.34 & 0.27 \\
        \midrule
        GPT-Image-1~\cite{gpt_image_1}                & 0.84 & 0.99 & 0.92 & 0.85 & 0.92 & 0.75 & 0.61 \\
        \midrule
        \multicolumn{8}{l}{\textit{Collaborative AI Systems}} \\
        ComfyAgent~\cite{xue2409comfybench}  & 0.32  & 0.69 & 0.30 & 0.33 & 0.50 & 0.04 & 0.04 \\
        \textbf{Ours}        & \textbf{0.90} & \textbf{1.00} & \textbf{1.00} & \textbf{1.00} & \textbf{0.97} & 0.62 & \textbf{0.80} \\
        \bottomrule
      \end{tabular}
    }
\end{table}

\subsubsection{Qualitative Results}

\begin{figure}[htbp]
  \centering
  \includegraphics[width=1.0\textwidth]{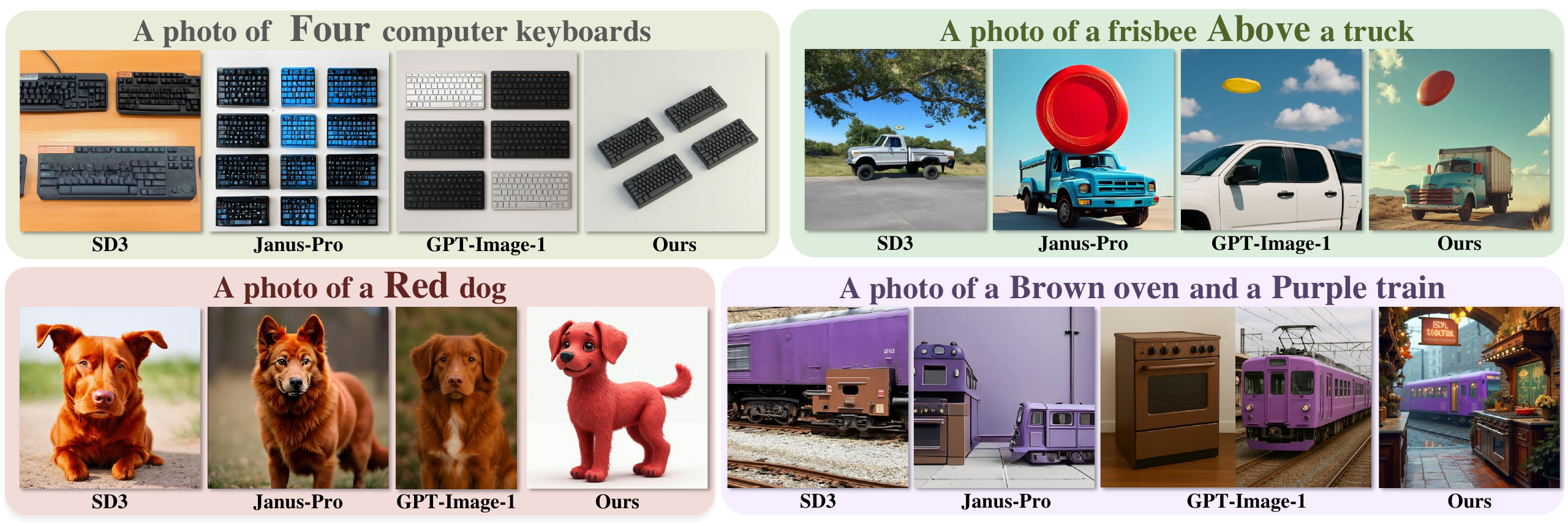}
  \vspace{-1.2em} 
  \caption{Qualitative comparison on challenging GenEval~\cite{ghosh2023geneval} cases.
Under constraints such as counting, color, position and attribute binding, only our method successfully satisfies all instructions, clearly outperforming SD3, Janus-Pro, and GPT-Image-1.}
  \label{fig:qualitative_geneval}
\end{figure}

Figure~\ref{fig:qualitative_geneval} showcases representative and challenging cases from GenEval. Our method follows prompts, outperforming existing models on core constraints like counting, color, position and Attribution Binding. In the counting task, only our system generates exactly four keyboards with clear visual separation. For atypical color and position, we demonstrate superior image quality and instruction consistency. Regarding attribute binding, models like SD3 and Janus-Pro often entangle attributes and fail to localize them correctly. GPT-Image-1, while generally instruction-following, often produces fragmented and visually incoherent compositions. In contrast, our method not only satisfies fine-grained directives but also integrates them into aesthetically coherent, contextually grounded scenes. These qualitative results corroborate the quantitative gains reported earlier.

\subsection{Image Editing}

\subsubsection{Quantitative Results}

We further evaluate our system’s image editing capability on the Reason‑Edit~\cite{huang2024smartedit}.
Following the setting in benchmark, we adopt the \textit{GPT‑score}~\cite{huang2024smartedit} as the evaluation metric. This score quantifies both the semantic fidelity to the editing instruction and the visual consistency of non-edited regions.

\begin{wrapfigure}[12]{r}{0.40\textwidth}
    \vspace{-1.0em} 
    \centering
    \setlength{\fboxsep}{0pt}
    \setlength{\fboxrule}{0pt}
    \includegraphics[width=\linewidth]{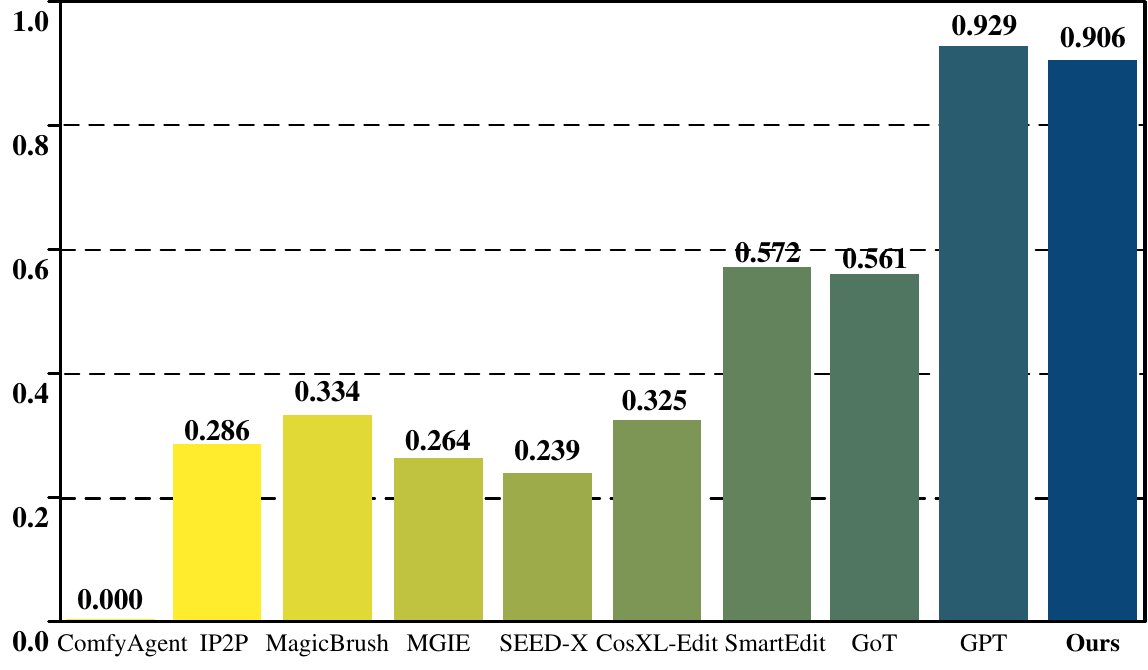}
    \vspace{-1.2em}
    \caption{Quantitative Comparison on Reason-Edit~\cite{huang2024smartedit} benchmark.}
    \label{fig:reason-edit_quan}
\end{wrapfigure}

We compare our method against the most advanced open-source baselines, including GoT~\cite{fang2025got}, SmartEdit~\cite{huang2024smartedit}, CosXL‑Edit~\cite{huang2024smartedit}, SEED-X~\cite{ge2024seed}, MGIE~\cite{fu2023guiding}, MagicBrush~\cite{zhang2023magicbrush} and IP2P~\cite{brooks2023instructpix2pix}, as well as the strongest closed-source model, GPT-Image-1.
As shown in Figure~\ref{fig:reason-edit_quan}, our method achieves a score of \textbf{0.906}—the highest among all open-source frameworks. This result represents a substantial improvement of \textbf{+0.334} over the previous open-source SOTA SmartEdit (0.572).

Moreover, our method achieves performance competitive with GPT-Image-1 (0.929), narrowing the gap between open and closed models.
This gain arises from our system’s planning and feedback mechanism, which enables it to synthesise and combine the most effective editing workflows contributed by the ComfyUI community.
Through reasoning and iterative correction, our agent can adaptively select diverse workflows, improving the stability and precision of edits across varied scenarios.
These results highlight the reasoning‑driven editing capability of our system, and suggest strong potential for future performance gains through integration with more powerful workflows and models.


\subsubsection{Qualitative Results}

\begin{figure}[htbp]
  \centering
  \includegraphics[width=1.0\textwidth]{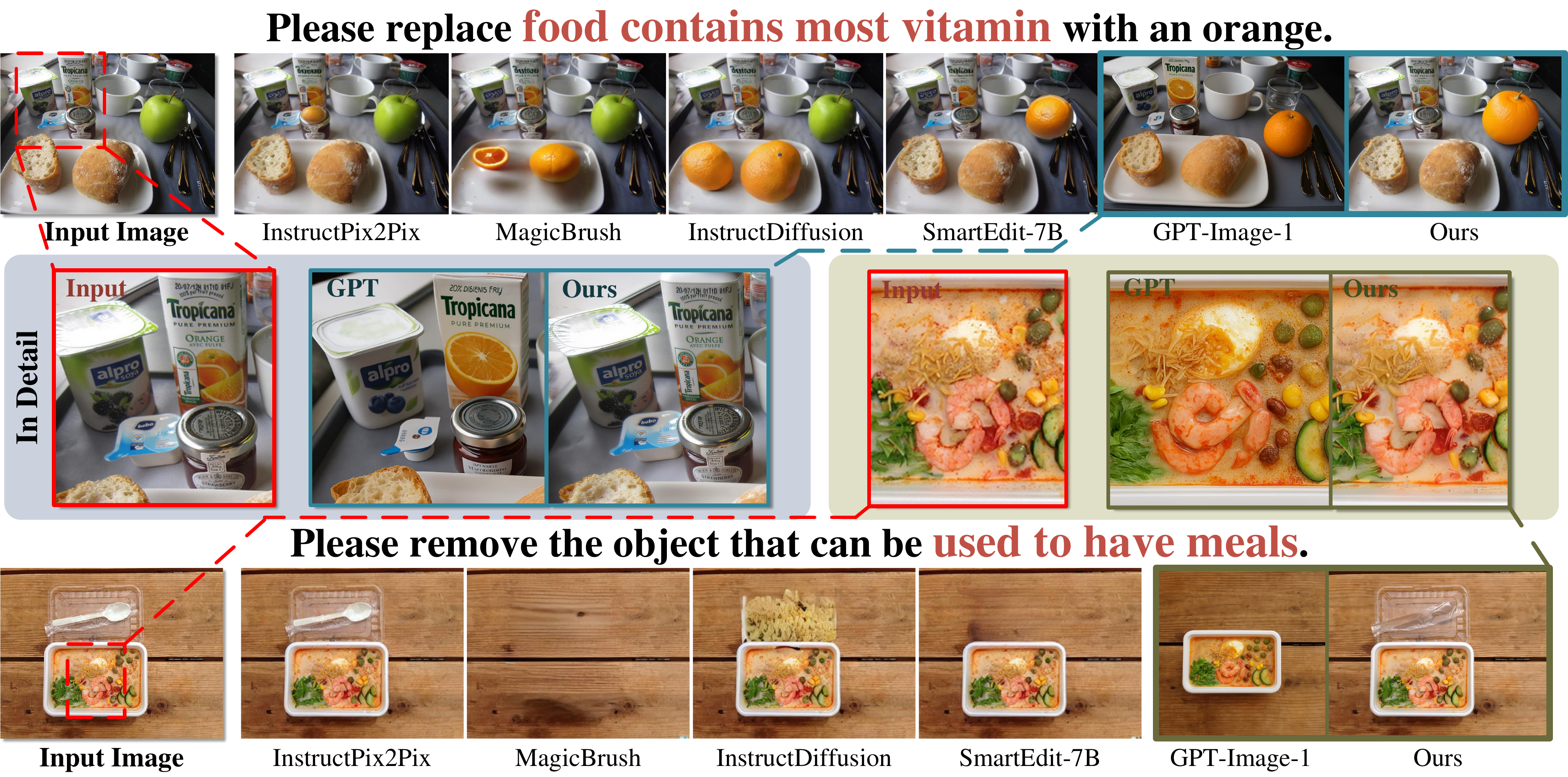}
  \vspace{-1.2em} 
  
  \caption{Qualitative Comparison on Reason-edit~\cite{huang2024smartedit} benchmark. }
  \label{fig:reason-edit_quali}
\end{figure}

We further present qualitative results to assess the semantic understanding and visual fidelity of our system under challenging editing instructions. We select two representative tasks from the Reason‑Edit~\cite{huang2024smartedit} benchmark. As shown in Figure~\ref{fig:reason-edit_quali}, our method consistently demonstrates the most faithful and visually coherent results across both tasks. Compared to existing open-source baselines, our system not only identifies the correct semantic target (e.g., apple vs. bread vs. orange juice) but also executes edits with minimal disruption to adjacent regions. 

Although GPT-Image-1 succeeds in executing editing instructions, it struggles to maintain visual consistency in non-edited regions. As illustrated in Figure~\ref{fig:reason-edit_quali}, GPT-Image-1 loses details in non-edited areas (e.g., patterns on the juice box, yogurt container, and jam jar in the zoom-in views), alters color tones and image style, inaccurately preserves materials (e.g., wood texture), and changes the original aspect ratio. These flaws are also mentioned in GPT-eval~\cite{yan2025gpt}.

In contrast, our method accomplishes the instructions with minimal edits, effectively preserving visual details, image style, material properties, and proportions. These observations highlight our method’s superior capability to perform precise and coherent edits.

\subsection{Ablation Study}

To isolate the contributions of our key design components, we conduct an ablation study on the ComfyBench benchmark (Table~\ref{tab:ablation}). We evaluate three variants: the full system, a version without search tree planning, and a version without feedback execution. Benefiting from our Semantic Workflow Interface, all variants achieve a 100\% pass rate; the main differences lie in the resolve rate.

\begin{table}[htbp]
  \centering
  \caption{Ablation results of ComfyMind with different architectures on ComfyBench.}
  \label{tab:ablation}
  \resizebox{\linewidth}{!}{
    \begin{tabular}{l|cccccccc}
      \toprule
      \multirow{2}{*}{\textbf{Agent}} & \multicolumn{2}{c}{\textbf{Vanilla}} & \multicolumn{2}{c}{\textbf{Complex}} & \multicolumn{2}{c}{\textbf{Creative}} & \multicolumn{2}{c}{\textbf{Total}} \\
      \cmidrule(lr){2-3} \cmidrule(lr){4-5} \cmidrule(lr){6-7} \cmidrule(lr){8-9}
      & \%Pass & \%Resolve & \%Pass & \%Resolve & \%Pass & \%Resolve & \%Pass & \%Resolve \\
      \midrule
      \textbf{Ours}       & \textbf{100.0} & \textbf{92.0} & \textbf{100.0} & \textbf{85.0} & \textbf{100.0} & \textbf{57.5} & \textbf{100.0} & \textbf{83.0} \\
      Ours w/o Tree Planning   & \textbf{100.0}  & 86.0  & \textbf{100.0}  & 43.4  & \textbf{100.0}  & 50.0  & \textbf{100.0}  & 66.0  \\      
      Ours w/o Feedback   & \textbf{100.0}  & 68.0  & \textbf{100.0}  & 55.0 
      & \textbf{100.0}  & 17.5   & \textbf{100.0}  & 54.5  \\    
      \bottomrule
    \end{tabular}
  }
\end{table}

Removing the search tree planning module results in a notable drop in task resolution, particularly on complex tasks (85.0\% to 43.4\%), underscoring its role in decomposing multi-step instructions and selecting suitable workflows. Similarly, disabling the local feedback mechanism significantly degrades performance, especially on creative tasks (57.5\% to 17.5\%), highlighting its importance for iterative correction and adaptive refinement. The results confirm that key components are essential for achieving high success rates in autonomous workflow construction.

\section{Conclusion} \label{sec:Conclusion}
In this work, we propose ComfyMind, a novel framework built on the ComfyUI platform that addresses key challenges in general-purpose visual generative AI. By conceptualizing visual content creation as a modular, semantically structured planning process and incorporating tree-based planning with local feedback execution, ComfyMind improves the stability and robustness of multi-stage workflows. Our framework outperforms previous open-source methods and achieves results comparable to GPT-Image-1 on benchmarks ComfyBench, GenEval, and Reason-Edit. ComfyMind offers a promising path towards scalable, open-source solutions for complex generative tasks.

\textbf{Discussion} This work focuses on developing a collaborative AI system with general-purpose visual generation capabilities, rather than generating complete workflow JSON files. The goal is to explore how such a system can plan, execute, and provide feedback for complex tasks through semantic planning, modular composition, and localized corrections. While we use ComfyUI workflow modules for execution, the workflow itself is a component of the system, not the research focus. By operating at a higher semantic level with the SWI, our approach reduces the complexity and uncertainty of workflow generation, offering a more robust and practical solution for real-world applications.

\small
\bibliography{main}

\begin{thebibliography}{53}
\providecommand{\natexlab}[1]{#1}
\providecommand{\url}[1]{\texttt{#1}}
\expandafter\ifx\csname urlstyle\endcsname\relax
  \providecommand{\doi}[1]{doi: #1}\else
  \providecommand{\doi}{doi: \begingroup \urlstyle{rm}\Url}\fi

\bibitem[Podell et~al.(2023)Podell, English, Lacey, Blattmann, Dockhorn, M{\"u}ller, Penna, and Rombach]{podell2023sdxl}
Dustin Podell, Zion English, Kyle Lacey, Andreas Blattmann, Tim Dockhorn, Jonas M{\"u}ller, Joe Penna, and Robin Rombach.
\newblock Sdxl: Improving latent diffusion models for high-resolution image synthesis.
\newblock \emph{arXiv preprint arXiv:2307.01952}, 2023.

\bibitem[Esser et~al.(2024)Esser, Kulal, Blattmann, Entezari, M{\"u}ller, Saini, Levi, Lorenz, Sauer, Boesel, et~al.]{esser2024scaling}
Patrick Esser, Sumith Kulal, Andreas Blattmann, Rahim Entezari, Jonas M{\"u}ller, Harry Saini, Yam Levi, Dominik Lorenz, Axel Sauer, Frederic Boesel, et~al.
\newblock Scaling rectified flow transformers for high-resolution image synthesis.
\newblock In \emph{Forty-first international conference on machine learning}, 2024.

\bibitem[Labs(2024{\natexlab{a}})]{blackforestlabs_flux_2024}
Black~Forest Labs.
\newblock Flux, 2024{\natexlab{a}}.
\newblock URL \url{https://github.com/black-forest-labs/flux}.

\bibitem[Brooks et~al.(2023)Brooks, Holynski, and Efros]{brooks2023instructpix2pix}
Tim Brooks, Aleksander Holynski, and Alexei~A Efros.
\newblock Instructpix2pix: Learning to follow image editing instructions.
\newblock In \emph{Proceedings of the IEEE/CVF conference on computer vision and pattern recognition}, pages 18392--18402, 2023.

\bibitem[Ai(2024)]{Stability_Ai_Cosxl}
Stability Ai.
\newblock Cosxl, 2024.
\newblock URL \url{https://huggingface.co/stabilityai/cosxl}.

\bibitem[Huang et~al.(2024)Huang, Xie, Wang, Yuan, Cun, Ge, Zhou, Dong, Huang, Zhang, et~al.]{huang2024smartedit}
Yuzhou Huang, Liangbin Xie, Xintao Wang, Ziyang Yuan, Xiaodong Cun, Yixiao Ge, Jiantao Zhou, Chao Dong, Rui Huang, Ruimao Zhang, et~al.
\newblock Smartedit: Exploring complex instruction-based image editing with multimodal large language models.
\newblock In \emph{Proceedings of the IEEE/CVF Conference on Computer Vision and Pattern Recognition}, pages 8362--8371, 2024.

\bibitem[HaCohen et~al.(2024)HaCohen, Chiprut, Brazowski, Shalem, Moshe, Richardson, Levin, Shiran, Zabari, Gordon, et~al.]{hacohen2024ltx}
Yoav HaCohen, Nisan Chiprut, Benny Brazowski, Daniel Shalem, Dudu Moshe, Eitan Richardson, Eran Levin, Guy Shiran, Nir Zabari, Ori Gordon, et~al.
\newblock Ltx-video: Realtime video latent diffusion.
\newblock \emph{arXiv preprint arXiv:2501.00103}, 2024.

\bibitem[Tencent(2024)]{hunyuanvideo}
Tencent.
\newblock Hunyuanvideo, 2024.
\newblock URL \url{https://github.com/Tencent/HunyuanVideo}.

\bibitem[Wang et~al.(2025)Wang, Ai, Wen, Mao, Xie, Chen, Yu, Zhao, Yang, Zeng, et~al.]{wang2025wan}
Ang Wang, Baole Ai, Bin Wen, Chaojie Mao, Chen-Wei Xie, Di~Chen, Feiwu Yu, Haiming Zhao, Jianxiao Yang, Jianyuan Zeng, et~al.
\newblock Wan: Open and advanced large-scale video generative models.
\newblock \emph{arXiv preprint arXiv:2503.20314}, 2025.

\bibitem[Ge et~al.(2024)Ge, Zhao, Zhu, Ge, Yi, Song, Li, Ding, and Shan]{ge2024seed}
Yuying Ge, Sijie Zhao, Jinguo Zhu, Yixiao Ge, Kun Yi, Lin Song, Chen Li, Xiaohan Ding, and Ying Shan.
\newblock Seed-x: Multimodal models with unified multi-granularity comprehension and generation.
\newblock \emph{arXiv preprint arXiv:2404.14396}, 2024.

\bibitem[Wu et~al.(2024{\natexlab{a}})Wu, Chen, Wu, Ma, Liu, Pan, Liu, Xie, Yu, Ruan, et~al.]{wu2024janus}
Chengyue Wu, Xiaokang Chen, Zhiyu Wu, Yiyang Ma, Xingchao Liu, Zizheng Pan, Wen Liu, Zhenda Xie, Xingkai Yu, Chong Ruan, et~al.
\newblock Janus: Decoupling visual encoding for unified multimodal understanding and generation.
\newblock \emph{arXiv preprint arXiv:2410.13848}, 2024{\natexlab{a}}.

\bibitem[Fang et~al.(2025)Fang, Duan, Wang, Huang, Li, Yan, Tian, Zeng, Zhao, Dai, et~al.]{fang2025got}
Rongyao Fang, Chengqi Duan, Kun Wang, Linjiang Huang, Hao Li, Shilin Yan, Hao Tian, Xingyu Zeng, Rui Zhao, Jifeng Dai, et~al.
\newblock Got: Unleashing reasoning capability of multimodal large language model for visual generation and editing.
\newblock \emph{arXiv preprint arXiv:2503.10639}, 2025.

\bibitem[OpenAI(2025)]{gpt_image_1}
OpenAI.
\newblock gpt-image-1, 2025.
\newblock URL \url{https://platform.openai.com/docs/models/gpt-image-1}.

\bibitem[Yan et~al.(2025)Yan, Ye, Li, Huang, Yuan, He, Lin, He, He, and Yuan]{yan2025gpt}
Zhiyuan Yan, Junyan Ye, Weijia Li, Zilong Huang, Shenghai Yuan, Xiangyang He, Kaiqing Lin, Jun He, Conghui He, and Li~Yuan.
\newblock Gpt-imgeval: A comprehensive benchmark for diagnosing gpt4o in image generation.
\newblock \emph{arXiv preprint arXiv:2504.02782}, 2025.

\bibitem[Gal et~al.(2024)Gal, Haviv, Alaluf, Bermano, Cohen-Or, and Chechik]{gal2024comfygen}
Rinon Gal, Adi Haviv, Yuval Alaluf, Amit~H Bermano, Daniel Cohen-Or, and Gal Chechik.
\newblock Comfygen: Prompt-adaptive workflows for text-to-image generation.
\newblock \emph{arXiv preprint arXiv:2410.01731}, 2024.

\bibitem[Xue et~al.(2024)Xue, Lu, Huang, Wang, Ouyang, and Bai]{xue2409comfybench}
Xiangyuan Xue, Zeyu Lu, Di~Huang, Zidong Wang, Wanli Ouyang, and Lei Bai.
\newblock Comfybench: Benchmarking llm-based agents in comfyui for autonomously designing collaborative ai systems.
\newblock \emph{URL https://arxiv. org/abs/2409.01392}, 2024.

\bibitem[{ComfyUI Contributors}(2023)]{comfyui2023}
{ComfyUI Contributors}.
\newblock {ComfyUI}: A powerful and modular stable-diffusion gui.
\newblock \url{https://github.com/comfyanonymous/ComfyUI}, 2023.
\newblock Accessed: 2025-05-14.

\bibitem[Lewis et~al.(2020)Lewis, Perez, Piktus, Petroni, Karpukhin, Goyal, K{\"u}ttler, Lewis, Yih, Rockt{\"a}schel, et~al.]{lewis2020retrieval}
Patrick Lewis, Ethan Perez, Aleksandra Piktus, Fabio Petroni, Vladimir Karpukhin, Naman Goyal, Heinrich K{\"u}ttler, Mike Lewis, Wen-tau Yih, Tim Rockt{\"a}schel, et~al.
\newblock Retrieval-augmented generation for knowledge-intensive nlp tasks.
\newblock \emph{Advances in neural information processing systems}, 33:\penalty0 9459--9474, 2020.

\bibitem[Zhang et~al.(2023{\natexlab{a}})Zhang, Rao, and Agrawala]{zhang2023adding}
Lvmin Zhang, Anyi Rao, and Maneesh Agrawala.
\newblock Adding conditional control to text-to-image diffusion models.
\newblock In \emph{Proceedings of the IEEE/CVF international conference on computer vision}, pages 3836--3847, 2023{\natexlab{a}}.

\bibitem[Ye et~al.(2023)Ye, Zhang, Liu, Han, and Yang]{ye2023ip}
Hu~Ye, Jun Zhang, Sibo Liu, Xiao Han, and Wei Yang.
\newblock Ip-adapter: Text compatible image prompt adapter for text-to-image diffusion models.
\newblock \emph{arXiv preprint arXiv:2308.06721}, 2023.

\bibitem[Gal et~al.(2022)Gal, Alaluf, Atzmon, Patashnik, Bermano, Chechik, and Cohen-Or]{gal2022image}
Rinon Gal, Yuval Alaluf, Yuval Atzmon, Or~Patashnik, Amit~H Bermano, Gal Chechik, and Daniel Cohen-Or.
\newblock An image is worth one word: Personalizing text-to-image generation using textual inversion.
\newblock \emph{arXiv preprint arXiv:2208.01618}, 2022.

\bibitem[Ma et~al.(2024)Ma, Liu, Chen, Liu, Wu, Wu, Pan, Xie, Zhang, Zhao, et~al.]{ma2024janusflow}
Yiyang Ma, Xingchao Liu, Xiaokang Chen, Wen Liu, Chengyue Wu, Zhiyu Wu, Zizheng Pan, Zhenda Xie, Haowei Zhang, Liang Zhao, et~al.
\newblock Janusflow: Harmonizing autoregression and rectified flow for unified multimodal understanding and generation.
\newblock \emph{arXiv preprint arXiv:2411.07975}, 2024.

\bibitem[Team(2024)]{team2024chameleon}
Chameleon Team.
\newblock Chameleon: Mixed-modal early-fusion foundation models.
\newblock \emph{arXiv preprint arXiv:2405.09818}, 2024.

\bibitem[Wang et~al.(2024)Wang, Zhang, Luo, Sun, Cui, Wang, Zhang, Wang, Li, Yu, et~al.]{wang2024emu3}
Xinlong Wang, Xiaosong Zhang, Zhengxiong Luo, Quan Sun, Yufeng Cui, Jinsheng Wang, Fan Zhang, Yueze Wang, Zhen Li, Qiying Yu, et~al.
\newblock Emu3: Next-token prediction is all you need.
\newblock \emph{arXiv preprint arXiv:2409.18869}, 2024.

\bibitem[Wu et~al.(2024{\natexlab{b}})Wu, Zhang, Chen, Tang, Li, Fang, Zhu, Xie, Yin, Yi, et~al.]{wu2024vila}
Yecheng Wu, Zhuoyang Zhang, Junyu Chen, Haotian Tang, Dacheng Li, Yunhao Fang, Ligeng Zhu, Enze Xie, Hongxu Yin, Li~Yi, et~al.
\newblock Vila-u: a unified foundation model integrating visual understanding and generation.
\newblock \emph{arXiv preprint arXiv:2409.04429}, 2024{\natexlab{b}}.

\bibitem[Xie et~al.(2024)Xie, Mao, Bai, Zhang, Wang, Lin, Gu, Chen, Yang, and Shou]{xie2024show}
Jinheng Xie, Weijia Mao, Zechen Bai, David~Junhao Zhang, Weihao Wang, Kevin~Qinghong Lin, Yuchao Gu, Zhijie Chen, Zhenheng Yang, and Mike~Zheng Shou.
\newblock Show-o: One single transformer to unify multimodal understanding and generation.
\newblock \emph{arXiv preprint arXiv:2408.12528}, 2024.

\bibitem[Zhao et~al.(2024)Zhao, Song, Wang, Feng, Ding, Sun, Xiao, and Wang]{zhao2024monoformer}
Chuyang Zhao, Yuxing Song, Wenhao Wang, Haocheng Feng, Errui Ding, Yifan Sun, Xinyan Xiao, and Jingdong Wang.
\newblock Monoformer: One transformer for both diffusion and autoregression.
\newblock \emph{arXiv preprint arXiv:2409.16280}, 2024.

\bibitem[Xiao et~al.(2024)Xiao, Wang, Zhou, Yuan, Xing, Yan, Li, Wang, Huang, and Liu]{xiao2024omnigen}
Shitao Xiao, Yueze Wang, Junjie Zhou, Huaying Yuan, Xingrun Xing, Ruiran Yan, Chaofan Li, Shuting Wang, Tiejun Huang, and Zheng Liu.
\newblock Omnigen: Unified image generation.
\newblock \emph{arXiv preprint arXiv:2409.11340}, 2024.

\bibitem[Chen et~al.(2025)Chen, Wu, Liu, Pan, Liu, Xie, Yu, and Ruan]{chen2025janus}
Xiaokang Chen, Zhiyu Wu, Xingchao Liu, Zizheng Pan, Wen Liu, Zhenda Xie, Xingkai Yu, and Chong Ruan.
\newblock Janus-pro: Unified multimodal understanding and generation with data and model scaling.
\newblock \emph{arXiv preprint arXiv:2501.17811}, 2025.

\bibitem[Richards(2023)]{richards2023autogpt}
Toran~Bruce Richards.
\newblock {AutoGPT}.
\newblock \url{https://github.com/Significant-Gravitas/AutoGPT}, 2023.

\bibitem[Wu et~al.(2023)Wu, Bansal, Zhang, Wu, Li, Zhu, Jiang, Zhang, Zhang, Liu, et~al.]{wu2023autogen}
Qingyun Wu, Gagan Bansal, Jieyu Zhang, Yiran Wu, Beibin Li, Erkang Zhu, Li~Jiang, Xiaoyun Zhang, Shaokun Zhang, Jiale Liu, et~al.
\newblock Autogen: Enabling next-gen llm applications via multi-agent conversation.
\newblock \emph{arXiv preprint arXiv:2308.08155}, 2023.

\bibitem[Xie et~al.(2023)Xie, Zhou, Cheng, Shi, Weng, Liu, Hua, Zhao, Liu, Liu, et~al.]{xie2023openagents}
Tianbao Xie, Fan Zhou, Zhoujun Cheng, Peng Shi, Luoxuan Weng, Yitao Liu, Toh~Jing Hua, Junning Zhao, Qian Liu, Che Liu, et~al.
\newblock Openagents: An open platform for language agents in the wild.
\newblock \emph{arXiv preprint arXiv:2310.10634}, 2023.

\bibitem[Yao et~al.(2023)Yao, Zhao, Yu, Du, Shafran, Narasimhan, and Cao]{yao2023react}
Shunyu Yao, Jeffrey Zhao, Dian Yu, Nan Du, Izhak Shafran, Karthik Narasimhan, and Yuan Cao.
\newblock React: Synergizing reasoning and acting in language models.
\newblock In \emph{International Conference on Learning Representations (ICLR)}, 2023.

\bibitem[Hong et~al.(2023)Hong, Zheng, Chen, Cheng, Wang, Zhang, Wang, Yau, Lin, Zhou, et~al.]{hong2023metagpt}
Sirui Hong, Xiawu Zheng, Jonathan Chen, Yuheng Cheng, Jinlin Wang, Ceyao Zhang, Zili Wang, Steven Ka~Shing Yau, Zijuan Lin, Liyang Zhou, et~al.
\newblock Metagpt: Meta programming for multi-agent collaborative framework.
\newblock \emph{arXiv preprint arXiv:2308.00352}, 3\penalty0 (4):\penalty0 6, 2023.

\bibitem[Labs(2024{\natexlab{b}})]{blackforestlabs_flux_redux_2024}
Black~Forest Labs.
\newblock Flux.1-redux-dev, 2024{\natexlab{b}}.
\newblock URL \url{https://huggingface.co/black-forest-labs/FLUX.1-Redux-dev}.

\bibitem[Mao et~al.(2025)Mao, Zhang, Pan, Jiang, Han, Liu, and Zhou]{mao2025ace++}
Chaojie Mao, Jingfeng Zhang, Yulin Pan, Zeyinzi Jiang, Zhen Han, Yu~Liu, and Jingren Zhou.
\newblock Ace++: Instruction-based image creation and editing via context-aware content filling.
\newblock \emph{arXiv preprint arXiv:2501.02487}, 2025.

\bibitem[Kirillov et~al.(2023)Kirillov, Mintun, Ravi, Mao, Rolland, Gustafson, Xiao, Whitehead, Berg, Lo, et~al.]{kirillov2023segment}
Alexander Kirillov, Eric Mintun, Nikhila Ravi, Hanzi Mao, Chloe Rolland, Laura Gustafson, Tete Xiao, Spencer Whitehead, Alexander~C Berg, Wan-Yen Lo, et~al.
\newblock Segment anything.
\newblock In \emph{2023 IEEE/CVF International Conference on Computer Vision (ICCV)}, pages 3992--4003. IEEE, 2023.

\bibitem[Ravi et~al.(2024)Ravi, Gabeur, Hu, Hu, Ryali, Ma, Khedr, R{\"a}dle, Rolland, Gustafson, et~al.]{ravi2024sam}
Nikhila Ravi, Valentin Gabeur, Yuan-Ting Hu, Ronghang Hu, Chaitanya Ryali, Tengyu Ma, Haitham Khedr, Roman R{\"a}dle, Chloe Rolland, Laura Gustafson, et~al.
\newblock Sam 2: Segment anything in images and videos.
\newblock \emph{arXiv preprint arXiv:2408.00714}, 2024.

\bibitem[Yao et~al.()Yao, Zhao, Yu, Du, Shafran, Narasimhan, and Cao]{yaoreact}
Shunyu Yao, Jeffrey Zhao, Dian Yu, Nan Du, Izhak Shafran, Karthik~R Narasimhan, and Yuan Cao.
\newblock React: Synergizing reasoning and acting in language models.
\newblock In \emph{The Eleventh International Conference on Learning Representations}.

\bibitem[Ghosh et~al.(2023)Ghosh, Hajishirzi, and Schmidt]{ghosh2023geneval}
Dhruba Ghosh, Hannaneh Hajishirzi, and Ludwig Schmidt.
\newblock Geneval: An object-focused framework for evaluating text-to-image alignment.
\newblock \emph{Advances in Neural Information Processing Systems}, 36:\penalty0 52132--52152, 2023.

\bibitem[Brown et~al.(2020)Brown, Mann, Ryder, Subbiah, Kaplan, Dhariwal, Neelakantan, Shyam, Sastry, Askell, et~al.]{brown2020language}
Tom Brown, Benjamin Mann, Nick Ryder, Melanie Subbiah, Jared~D Kaplan, Prafulla Dhariwal, Arvind Neelakantan, Pranav Shyam, Girish Sastry, Amanda Askell, et~al.
\newblock Language models are few-shot learners.
\newblock \emph{Advances in neural information processing systems}, 33:\penalty0 1877--1901, 2020.

\bibitem[Wei et~al.(2022)Wei, Wang, Schuurmans, Bosma, Xia, Chi, Le, Zhou, et~al.]{wei2022chain}
Jason Wei, Xuezhi Wang, Dale Schuurmans, Maarten Bosma, Fei Xia, Ed~Chi, Quoc~V Le, Denny Zhou, et~al.
\newblock Chain-of-thought prompting elicits reasoning in large language models.
\newblock \emph{Advances in neural information processing systems}, 35:\penalty0 24824--24837, 2022.

\bibitem[Wang et~al.(2022)Wang, Wei, Schuurmans, Le, Chi, Narang, Chowdhery, and Zhou]{wang2022self}
Xuezhi Wang, Jason Wei, Dale Schuurmans, Quoc Le, Ed~Chi, Sharan Narang, Aakanksha Chowdhery, and Denny Zhou.
\newblock Self-consistency improves chain of thought reasoning in language models.
\newblock \emph{arXiv preprint arXiv:2203.11171}, 2022.

\bibitem[Rombach et~al.(2022)Rombach, Blattmann, Lorenz, Esser, and Ommer]{rombach2022high}
Robin Rombach, Andreas Blattmann, Dominik Lorenz, Patrick Esser, and Bj{\"o}rn Ommer.
\newblock High-resolution image synthesis with latent diffusion models.
\newblock In \emph{Proceedings of the IEEE/CVF conference on computer vision and pattern recognition}, pages 10684--10695, 2022.

\bibitem[Ramesh et~al.(2022)Ramesh, Dhariwal, Nichol, Chu, and Chen]{ramesh2022hierarchical}
Aditya Ramesh, Prafulla Dhariwal, Alex Nichol, Casey Chu, and Mark Chen.
\newblock Hierarchical text-conditional image generation with clip latents.
\newblock \emph{arXiv preprint arXiv:2204.06125}, 1\penalty0 (2):\penalty0 3, 2022.

\bibitem[Sun et~al.(2024)Sun, Jiang, Chen, Zhang, Peng, Luo, and Yuan]{sun2024autoregressive}
Peize Sun, Yi~Jiang, Shoufa Chen, Shilong Zhang, Bingyue Peng, Ping Luo, and Zehuan Yuan.
\newblock Autoregressive model beats diffusion: Llama for scalable image generation.
\newblock \emph{arXiv preprint arXiv:2406.06525}, 2024.

\bibitem[Liu et~al.(2024)Liu, Yan, Zaharia, and Abbeel]{liu2024world}
Hao Liu, Wilson Yan, Matei Zaharia, and Pieter Abbeel.
\newblock World model on million-length video and language with ringattention.
\newblock \emph{arXiv e-prints}, pages arXiv--2402, 2024.

\bibitem[Fu et~al.(2023)Fu, Hu, Du, Wang, Yang, and Gan]{fu2023guiding}
Tsu-Jui Fu, Wenze Hu, Xianzhi Du, William~Yang Wang, Yinfei Yang, and Zhe Gan.
\newblock Guiding instruction-based image editing via multimodal large language models.
\newblock \emph{arXiv preprint arXiv:2309.17102}, 2023.

\bibitem[Zhang et~al.(2023{\natexlab{b}})Zhang, Mo, Chen, Sun, and Su]{zhang2023magicbrush}
Kai Zhang, Lingbo Mo, Wenhu Chen, Huan Sun, and Yu~Su.
\newblock Magicbrush: A manually annotated dataset for instruction-guided image editing.
\newblock \emph{Advances in Neural Information Processing Systems}, 36:\penalty0 31428--31449, 2023{\natexlab{b}}.

\bibitem[Niu et~al.(2025)Niu, Ning, Zheng, Lin, Jin, Liao, Ning, Zhu, and Yuan]{niu2025wise}
Yuwei Niu, Munan Ning, Mengren Zheng, Bin Lin, Peng Jin, Jiaqi Liao, Kunpeng Ning, Bin Zhu, and Li~Yuan.
\newblock Wise: A world knowledge-informed semantic evaluation for text-to-image generation.
\newblock \emph{arXiv preprint arXiv:2503.07265}, 2025.

\bibitem[Chen et~al.(2023)Chen, Yu, Ge, Yao, Xie, Wu, Wang, Kwok, Luo, Lu, et~al.]{chen2023pixart}
Junsong Chen, Jincheng Yu, Chongjian Ge, Lewei Yao, Enze Xie, Yue Wu, Zhongdao Wang, James Kwok, Ping Luo, Huchuan Lu, et~al.
\newblock Fast training of diffusion transformer for photorealistic text-to-image synthesis.
\newblock \emph{arXiv preprint arXiv:2310.00426}, 2023.

\bibitem[Li et~al.(2024)Li, Kamko, Akhgari, Sabet, Xu, and Doshi]{li2024playground}
Daiqing Li, Aleks Kamko, Ehsan Akhgari, Ali Sabet, Linmiao Xu, and Suhail Doshi.
\newblock Playground v2. 5: Three insights towards enhancing aesthetic quality in text-to-image generation.
\newblock \emph{arXiv preprint arXiv:2402.17245}, 2024.

\bibitem[Kou et~al.(2024)Kou, Jin, Liu, Liu, Ma, Jia, Chen, Jiang, and Deng]{kou2024orthus}
Siqi Kou, Jiachun Jin, Zhihong Liu, Chang Liu, Ye~Ma, Jian Jia, Quan Chen, Peng Jiang, and Zhijie Deng.
\newblock Orthus: Autoregressive interleaved image-text generation with modality-specific heads.
\newblock \emph{arXiv preprint arXiv:2412.00127}, 2024.

\end{thebibliography}
\bibliographystyle{unsrtnat}

\newpage
\appendix

\appendix

\section{Supplementary Experiments} \label{sec:Supplementary_Experiments}

\subsection{Supplementary Ablation Study}

\begin{table}[htbp]
  \centering
  \caption{Ablation results of ComfyMind with different LLMs.}
  \label{tab:sup_ablation}
  \resizebox{\linewidth}{!}{
    \begin{tabular}{l|cccccccc}
      \toprule
      \multirow{2}{*}{\textbf{Agent}} & \multicolumn{2}{c}{\textbf{Vanilla}} & \multicolumn{2}{c}{\textbf{Complex}} & \multicolumn{2}{c}{\textbf{Creative}} & \multicolumn{2}{c}{\textbf{Total}} \\
      \cmidrule(lr){2-3} \cmidrule(lr){4-5} \cmidrule(lr){6-7} \cmidrule(lr){8-9}
      & \%Pass & \%Resolve & \%Pass & \%Resolve & \%Pass & \%Resolve & \%Pass & \%Resolve \\
      \midrule
      ComfyMind with GPT-4o       & \textbf{100.0} & \textbf{92.0} & \textbf{100.0} & \textbf{85.0} & \textbf{100.0} & 57.5 & \textbf{100.0} & \textbf{83.0} \\
      ComfyMind with Deepseek-V3   & \textbf{100.0}  & 90.0  & \textbf{100.0}  & 71.7  & \textbf{100.0}  & \textbf{60.0}  & \textbf{100.0}  & 78.5  \\         
      \bottomrule
    \end{tabular}
  }
\end{table}

To further demonstrate the robustness of our system, we conducted evaluations on the ComfyBench benchmark using different LLMs in ComfyMind. As shown in Table~\ref{tab:sup_ablation}, both Deepseek-V3 and GPT-4o achieve strong performance when employed as the primary LLMs. Specifically, both models reached a 100\% task pass rate and approximately 80\% overall task completion rate. These results further confirm the stability and reliability of our system across different underlying LLMs.

\subsection{World knowledge-Informed Semantic Synthesis}

\begin{table}[htbp]
  \centering
  \caption{Evaluation of World Knowledge-Informed Semantic Synthesis on WISE~\cite{niu2025wise} Benchmark.}
  \label{tab:t2i-wise}
  \resizebox{\linewidth}{!}{
  \begin{tabular}{l|cccccc|c}
    \toprule
    \textbf{Method} & \textbf{Cultural} & \textbf{Time} & \textbf{Space} & \textbf{Biology} & \textbf{Physics} & \textbf{Chemistry} & \textbf{Overall}\\
    \midrule
    \multicolumn{8}{l}{\textit{Dedicated T2I Models}} \\
    FLUX.1-dev~\cite{blackforestlabs_flux_2024} & 0.48 & 0.58 & 0.62 & 0.42 & 0.51 & 0.35 & 0.50 \\
    FLUX.1-schnell~\cite{blackforestlabs_flux_2024} & 0.39 & 0.44 & 0.50 & 0.31 & 0.44 & 0.26 & 0.40 \\
    PixArt-Alpha~\cite{chen2023pixart} & 0.45 & 0.50 & 0.48 & 0.49 & 0.56 & 0.34 & 0.47 \\
    playground-v2.5~\cite{li2024playground} & 0.49 & 0.58 & 0.55 & 0.43 & 0.48 & 0.33 & 0.49 \\
    SDv1.5~\cite{rombach2022high} & 0.34 & 0.35 & 0.32 & 0.28 & 0.29 & 0.21 & 0.32 \\
    SDv2.1~\cite{rombach2022high} & 0.30 & 0.38 & 0.35 & 0.33 & 0.34 & 0.21 & 0.32 \\
    SD-XL-base-0.9~\cite{podell2023sdxl} & 0.43 & 0.48 & 0.47 & 0.44 & 0.45 & 0.27 & 0.43 \\
    SD3-Medium~\cite{esser2024scaling} & 0.42 & 0.44 & 0.48 & 0.39 & 0.47 & 0.29 & 0.42 \\
    SD3.5-Medium~\cite{esser2024scaling} & 0.43 & 0.50 & 0.52 & 0.41 & 0.53 & 0.33 & 0.45 \\
    SD3.5-Large~\cite{esser2024scaling} & 0.44 & 0.50 & 0.58 & 0.44 & 0.52 & 0.31 & 0.46 \\
    \midrule
    \multicolumn{8}{l}{\textit{Unify MLLM Models}} \\
    Emu3~\cite{wang2024emu3} & 0.34 & 0.45 & 0.48 & 0.41 & 0.45 & 0.27 & 0.39 \\
    Janus-1.3B~\cite{wu2024janus} & 0.16 & 0.26 & 0.35 & 0.28 & 0.30 & 0.14 & 0.23 \\
    JanusFlow-1.3B~\cite{ma2024janusflow} & 0.13 & 0.26 & 0.28 & 0.20 & 0.19 & 0.11 & 0.18 \\
    Janus-Pro-1B~\cite{chen2025janus} & 0.20 & 0.28 & 0.45 & 0.24 & 0.32 & 0.16 & 0.26 \\
    Janus-Pro-7B~\cite{chen2025janus} & 0.30 & 0.37 & 0.49 & 0.36 & 0.42 & 0.26 & 0.35 \\
    Orthus-7B-base~\cite{kou2024orthus} & 0.07 & 0.10 & 0.12 & 0.15 & 0.15 & 0.10 & 0.10 \\
    Orthus-7B-instruct~\cite{kou2024orthus} & 0.23 & 0.31 & 0.38 & 0.28 & 0.31 & 0.20 & 0.27 \\
    show-o-demo~\cite{xie2024show} & 0.28 & 0.36 & 0.40 & 0.23 & 0.33 & 0.22 & 0.30 \\
    show-o-demo-512~\cite{xie2024show} & 0.28 & 0.40 & 0.48 & 0.30 & 0.46 & 0.30 & 0.35 \\
    vila-u-7b-256~\cite{wu2024vila} & 0.26 & 0.33 & 0.37 & 0.35 & 0.39 & 0.23 & 0.31 \\
    \midrule
    GPT-Image-1~\cite{gpt_image_1} & 0.81 & 0.71 & 0.89 & \textbf{0.83} & \textbf{0.79} & 0.74 & 0.80 \\
    \midrule
    \multicolumn{8}{l}{\textit{Collaborative AI Systems}} \\
    \textbf{Ours} & \textbf{0.90} & \textbf{0.79} & \textbf{0.92} & 0.77 & \textbf{0.79} & \textbf{0.82} & \textbf{0.85} \\
    \bottomrule
  \end{tabular}
  }
\end{table}

To assess our system's capabilities in complex semantic understanding, reasoning, and world knowledge integration for text-to-image generation, we conduct evaluation on the recent WISE~\cite{niu2025wise} benchmark. This benchmark contains three primary categories: cultural commonsense, spatiotemporal reasoning (including Space and Time subcategories), and natural sciences (comprising Physics, Chemistry, and Biology subfields), totaling 25 specialized domains with 1,000 challenging prompts.

The evaluation metric WiScore combines Consistency, Realism, and Aesthetic Quality through weighted normalization, with a maximum score of 1. Higher WiScore indicates stronger capability in accurately depicting objects and concepts using world knowledge.      As shown in Table~\ref{tab:t2i-wise}, \textbf{our method achieves a superior score of 0.85, surpassing all models, including GPT-Image-1 (0.80)}.      Our approach significantly enhances world knowledge integration for open-source solutions, outperforming FLUX.1-dev (0.50) by 0.35 points and enabling open-source models to match GPT-Image-1's performance.   The exceptional performance on WISE confirms our system's generalizability and high-quality output in generative tasks.

\section{Additional Results and Generation Examples} \label{sec:Additional_Examples}

\begin{figure}[htbp]
  \centering
  \includegraphics[width=1.0\textwidth]{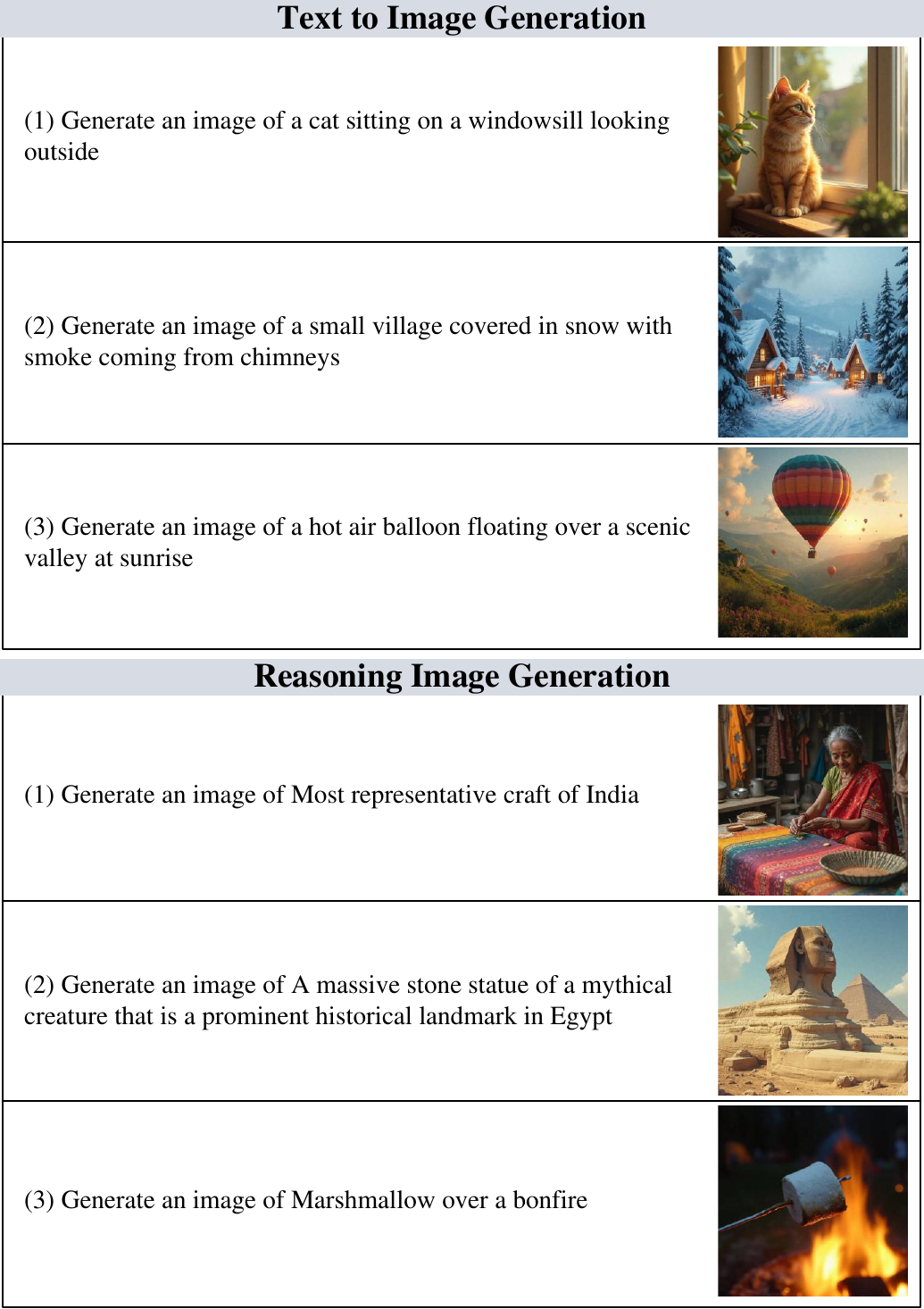}
  \caption{More examples generated by ComfyMind}
  \label{fig:appshow_1}
\end{figure}

\begin{figure}[htbp]
  \centering
  \includegraphics[width=1.0\textwidth]{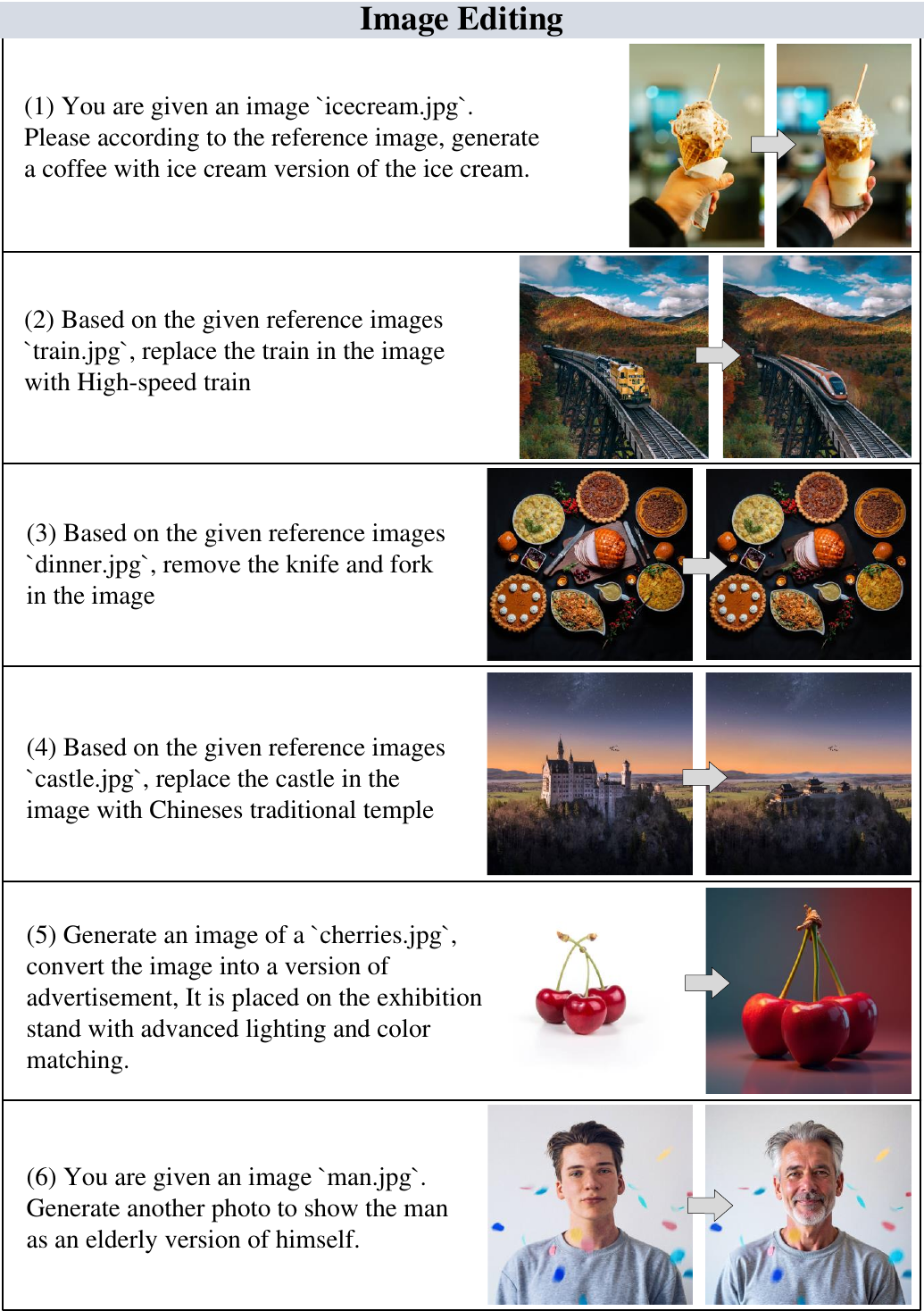}
  \caption{More examples generated by ComfyMind}
  \label{fig:appshow_2}
\end{figure}

\begin{figure}[htbp]
  \centering
  \includegraphics[width=1.0\textwidth]{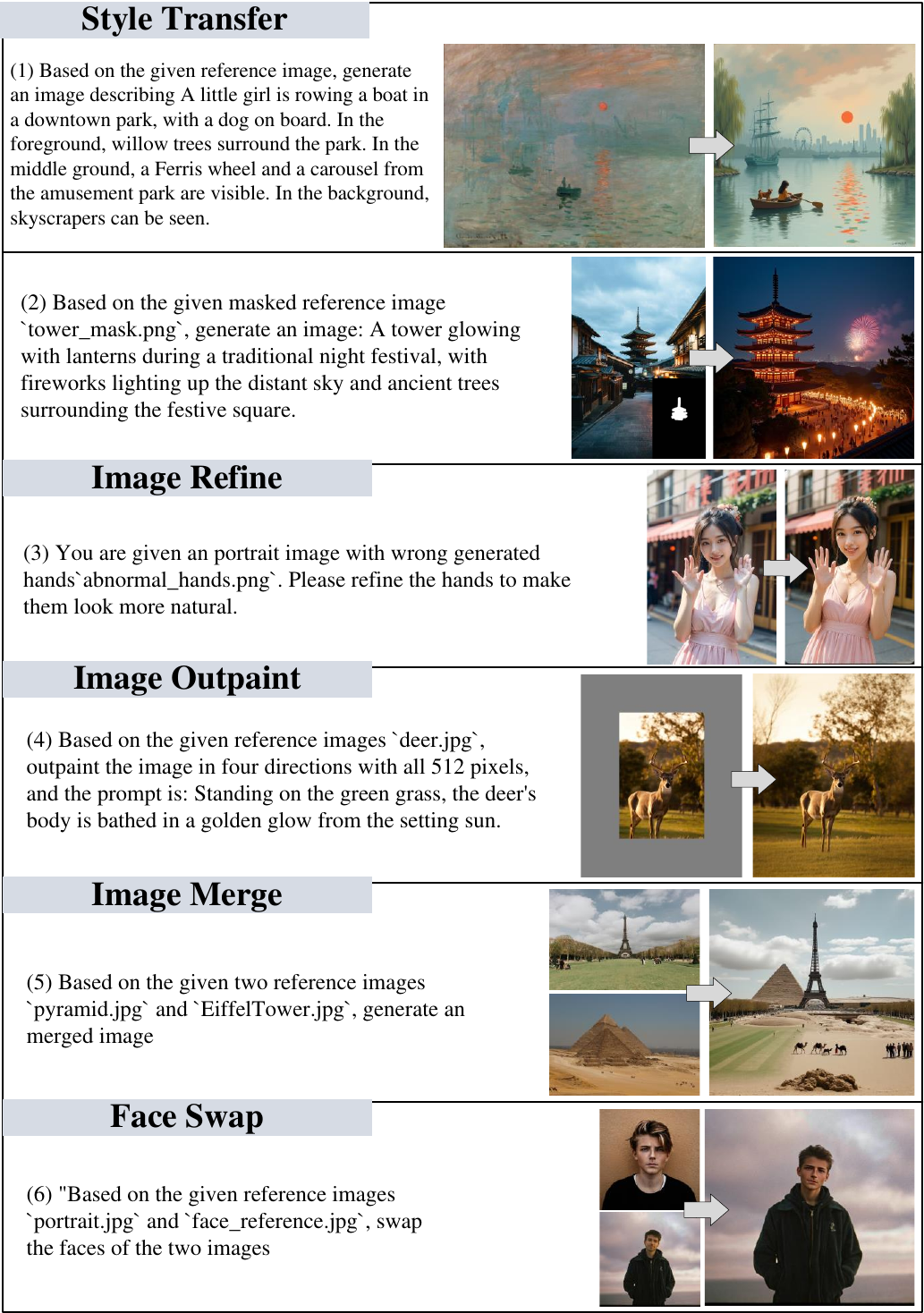}
  \caption{More examples generated by ComfyMind}
  \label{fig:appshow_3}
\end{figure}

As an extended demonstration, we present ComfyMind's exemplary outputs across diverse generative tasks in Figures~\ref{fig:appshow_1}, ~\ref{fig:appshow_2}, and ~\ref{fig:appshow_3}, substantiating its versatile adaptability and superior performance in cross-domain generation scenarios.

\section{Implementation Details} \label{sec:Implementation_Details}

\subsection{System Architecture} 
Logically, the system is divided into two independent subsystems: the ComfyUI server and the automation control system. Both are deployed on the same physical server but operate within isolated runtime environments and listen on separate ports. This design ensures robust separation and scalability in task concurrency handling and resource management. The server is equipped with an NVIDIA RTX A6000 GPU with 48GB of VRAM, providing sufficient computational capacity for generation tasks.

Within the system architecture, the ComfyUI platform is responsible for executing atomic workflows. It is deployed as a local service that listens on a designated port and provides dual interaction interfaces. On one hand, it exposes a web interface through which users can access the frontend via a browser to edit nodes, configure parameters, and perform real-time executions. On the other hand, ComfyUI incorporates both HTTP and WebSocket protocols, supporting standardized RESTful API task submissions and result retrieval, while enabling real-time progress tracking via WebSocket. In this study, the latter interface is primarily utilized. Technically, the client loads an adapted workflow in JSON format, submits a generation request via the HTTP API, and establishes a persistent WebSocket connection to receive real-time execution updates and final results. Built atop the ComfyUI execution layer, ComfyMind functions as the high-level planning and task management system, responsible for process orchestration, scheduling, and result evaluation. It is developed based on the ComfyAgent framework.

\subsection{Compilation of Atomic Workflow Library and Descriptive Documents}
To ensure adaptability across multimodal generation and editing tasks, a diverse set of atomic workflows was systematically collected, curated, and organized to form a foundational resource for general-purpose content generation. These workflows were sourced through a combination of selective acquisition and custom development, aiming to balance stability and broad applicability.

First, standard workflows from the ComfyBench benchmark suite were systematically tested. Suboptimal examples were discarded to ensure overall quality. Second, a curated selection of high-quality workflows was extracted from the official ComfyUI website and documentation to meet common generation needs. Additionally, workflows with strong user ratings and proven effectiveness were sourced from popular communities such as Civitai and OpenArt, enhancing the practicality and robustness of the workflow library.

Beyond these standardized resources, functional workflows tailored for complex reasoning tasks were developed by leveraging the flexibility of ComfyMind, in conjunction with LLMs and VLMs. These workflows include capabilities for reason generation, reason editing, and prompt enhancement, supplementing limitations in the standard resource base.

\subsection{Supplementary Description of the ComfyMind Method}

\subsubsection{Semantic Workflow Interface Details}
To enable SWI, a unified semantic description schema was developed for all atomic workflows. The system only requires a single, consistent atomic workflow description document to support all reasoning tasks. This document employs a hybrid format combining natural language and standardized parameter interfaces to specify each workflow’s core functionality, input/output requirements, and optional extensions.

The core structure includes: the required workflow name, the corresponding prompt and visual input. Optional components include supplementary semantic constraints such as resolution, frame rate, or upscaling ratio. This modular, uniform documentation enables semantic-level workflow selection and parameter matching during reasoning, without requiring parsing of underlying node structures—greatly improving planning efficiency and ease of expansion or maintenance.

\subsubsection{Execution Agent Details}
The execution agent in the ComfyMind system plays a crucial role in bridging high-level planning and low-level execution. Once the planning agent completes the decomposition of a subtask and outputs a set of invocation instructions, the execution agent is activated to handle parameter processing, workflow adaptation, execution, and result feedback. The overall logic adheres to the fundamental principle: "if execution succeeds, proceed to the next subtask; if it fails, return to the current node for local backtracking." This ensures tight integration between planning and execution.

Before initiating the execution process, certain specialized semantic workflow interface functions may require a preprocessing step. In such cases, the execution agent invokes LLMs to perform deep reasoning and semantic expansion on the user’s input prompt or initial task description. For example, if the user provides a command like “light passes through a prism,” the system leverages the LLM’s knowledge to infer related phenomena such as spectral dispersion and colorful light bands, thereby optimizing the prompt and providing richer and more accurate input conditions for the subsequent generation process.

Next, the execution agent parses the invocation parameters passed down from the planning agent, loads the corresponding atomic workflow template, and fills in placeholders with information such as the prompt text and input visual data paths. This results in a fully instantiated and executable workflow. To better adapt to the task’s specific requirements, the execution agent further adjusts workflow parameters, such as image resolution, sampling steps, and output frame rate. This adaptive tuning is performed at the parameter level without altering the underlying DAG structure of the atomic workflow, thus preserving the correctness of node connections and ensuring execution stability.

Once adaptation is complete, the execution agent submits the workflow to the ComfyUI platform through a standardized interface and establishes a real-time connection to monitor generation progress and status updates. After task execution finishes, the system automatically parses the generated results. It uses a visual-language understanding module to extract semantic descriptions, detailed information, and scene characteristics from the outputs, synchronizing them with the current workspace state. This update provides the most recent context for subsequent subtask reasoning, maintaining continuity and coherence in the reasoning chain.

If the entire process completes successfully, the execution agent returns a success signal and hands off the updated workspace state to the next planning node. If any exceptions occur during execution—such as parameter mismatches, insufficient generation quality, or timeouts—the execution agent returns a detailed failure reason and log information. This allows the planning agent to perform localized search backtracking and re-planning. Through this design, the execution agent effectively ensures seamless integration from high-level semantic instructions to concrete execution, while offering strong adaptability and robust error handling, significantly enhancing the system's overall stability and reasoning flexibility.

\subsubsection{Evaluation Agent Details}
The Evaluation Agent is responsible for assessing the quality of generated results and their consistency with the given instructions, ensuring that system outputs meet expected requirements. When the overall process concludes, the system invokes the Evaluation Agent to inspect the generated content against a unified standard and relay the results back to the higher-level planning module.

In its design, the Evaluation Agent adopts the classification-based assessment strategy used by ComfyAgent, dividing workflows according to task types such as text-to-image, image-to-image, text-to-video, image-to-video, and video-to-video, with tailored evaluation criteria for each category. Evaluation is conducted along two main dimensions: the quality of the generation—such as clarity, richness of detail, and visual naturalness—and adherence to instructions, verifying whether the output accurately reflects the user’s intended theme, setting, or stylistic directives. Each evaluation returns two components: a binary judgment indicating whether the current task is considered complete, which guides the system on whether to proceed or backtrack; and a detailed failure analysis when evaluation is unsuccessful, explicitly identifying reasons such as lack of detail, semantic deviation, or stylistic inconsistency, thereby aiding the Planning Agent in refining subsequent inference steps.

To accommodate varying tasks and application requirements, the Evaluation Agent supports dynamic threshold adjustment. The system can raise or lower evaluation standards depending on task complexity, quality expectations, or usage scenarios, enabling a flexible balance between content quality and execution efficiency.

\section{Limitation} \label{sec:Extended_Discussion}

Although ComfyMind supports modular workflow composition and automated planning, the current system lacks a user-friendly interface for manually customizing or modifying the sequence of atomic workflow invocations. Users have limited ability to adjust planning strategies, override intermediate steps, or specify task-specific preferences through the UI. This may hinder broader adoption among non-technical users or practitioners with domain-specific needs. Enhancing the interface to support more flexible and user-controllable planning customization is an important direction for future development.

\section{System Prompt} \label{sec:System_Prompt}
\begin{figure}[htbp]
  \centering
  \includegraphics[width=1.0\textwidth]{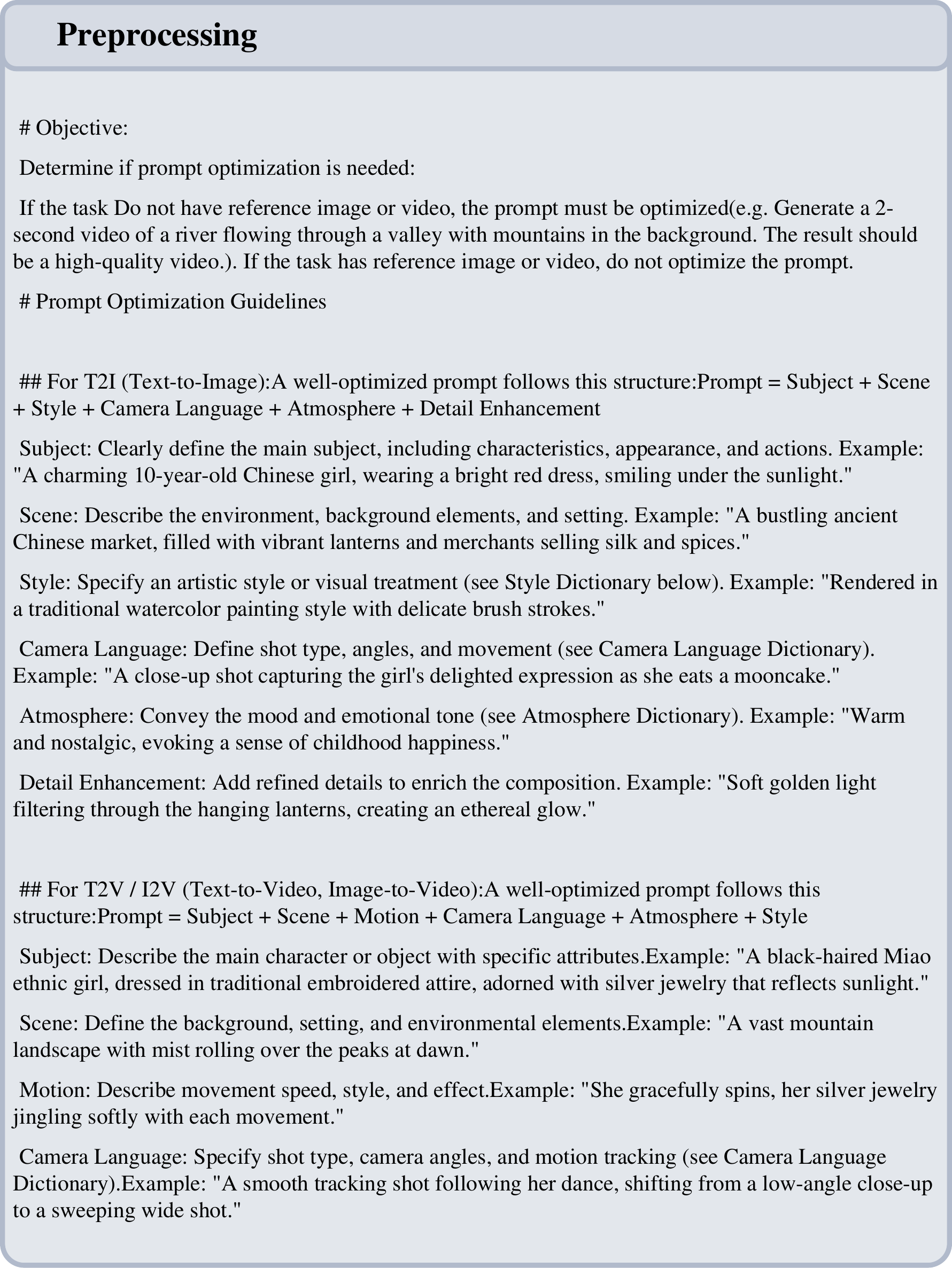}
  \caption{System Prompt for Preprocessing, Part 1}
  \label{fig:prompt1}
\end{figure}

\begin{figure}[htbp]
  \centering
  \includegraphics[width=1.0\textwidth]{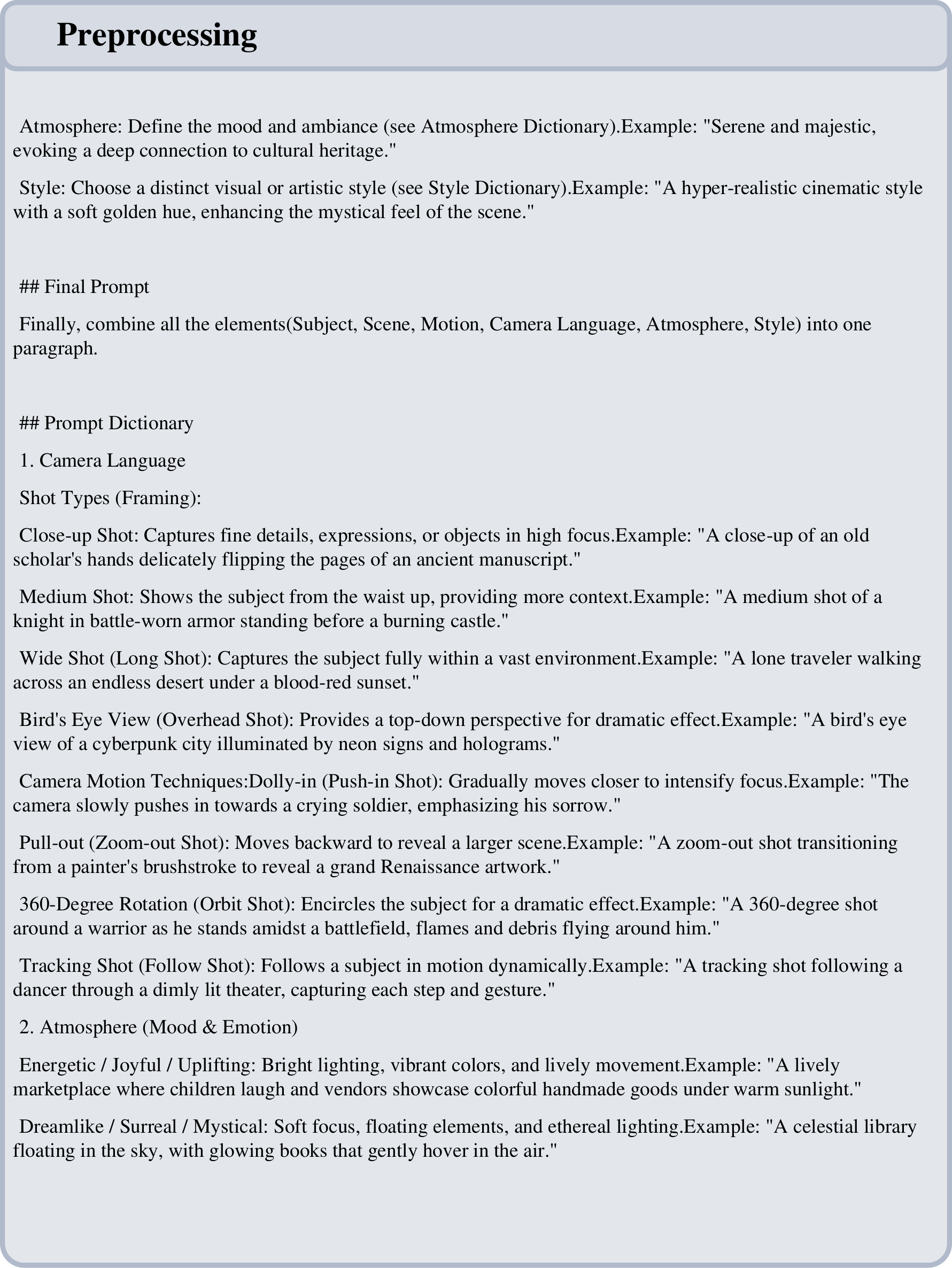}
  \caption{System Prompt for Preprocessing, Part 2}
  \label{fig:prompt2}
\end{figure}

\begin{figure}[htbp]
  \centering
  \includegraphics[width=1.0\textwidth]{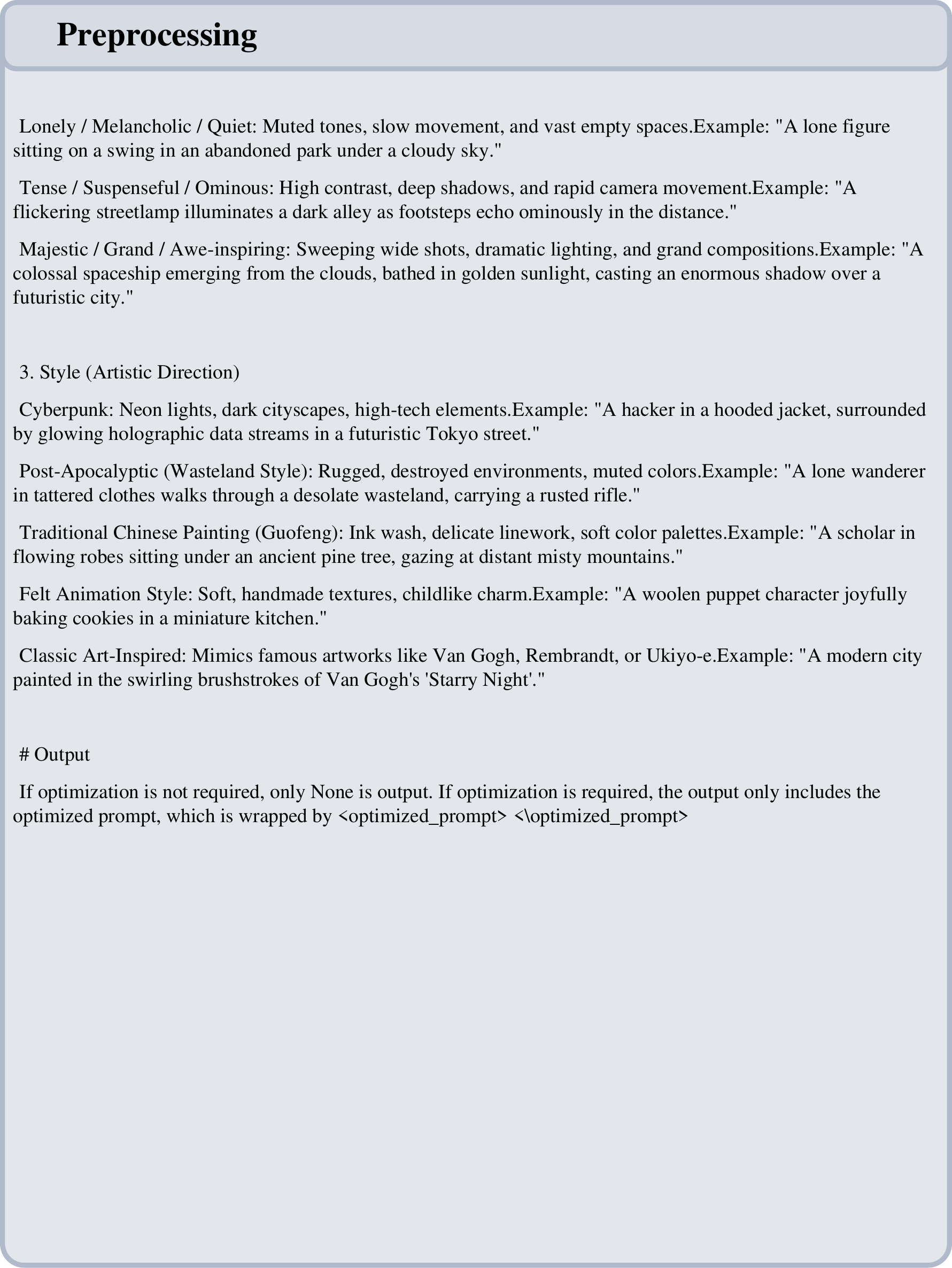}
  \caption{System Prompt for Preprocessing, Part 3}
  \label{fig:prompt3}
\end{figure}

\begin{figure}[htbp]
  \centering
  \includegraphics[width=1.0\textwidth]{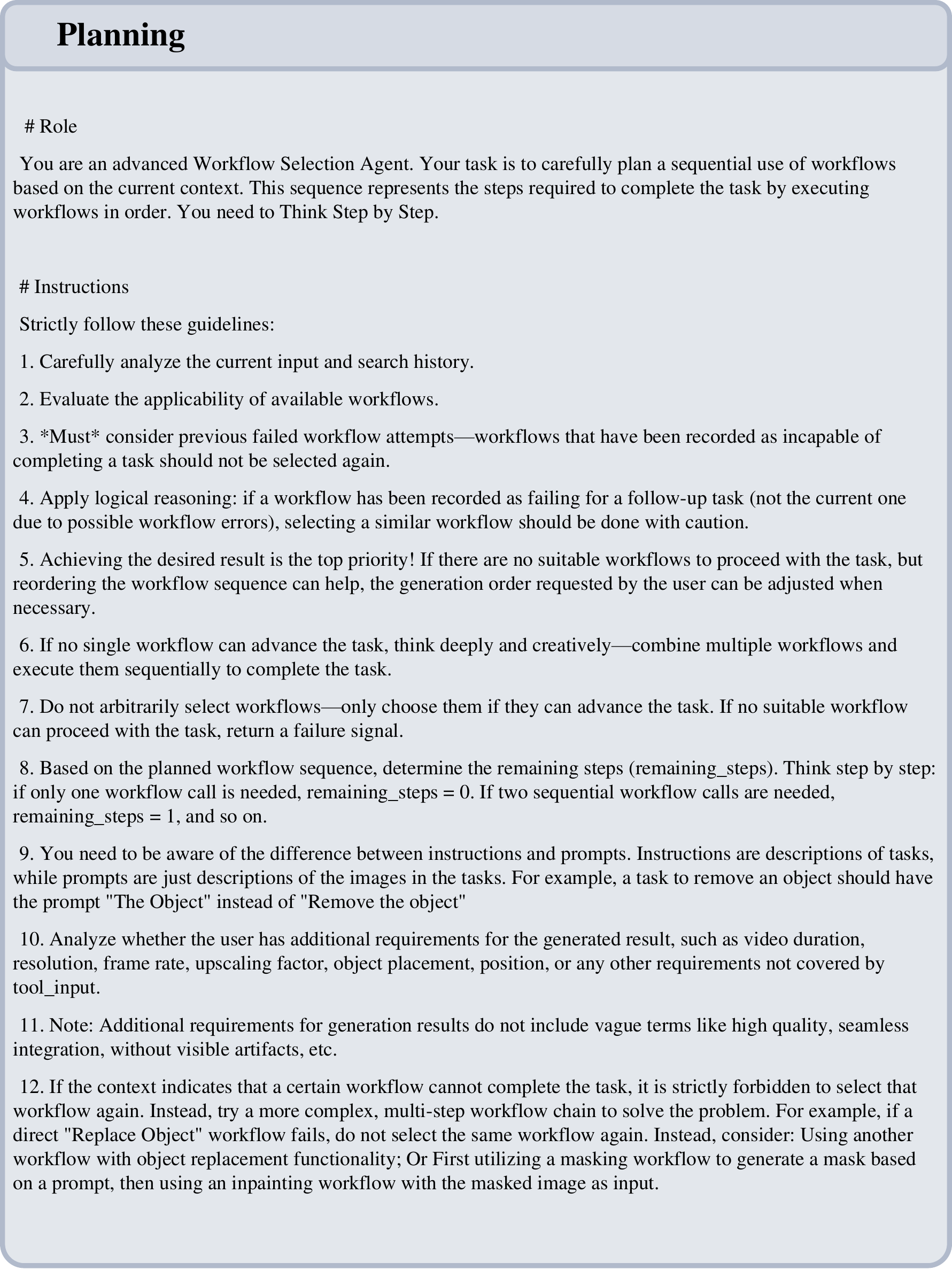}
  \caption{System Prompt for Planning, Part 1}
  \label{fig:prompt4}
\end{figure}

\begin{figure}[htbp]
  \centering
  \includegraphics[width=1.0\textwidth]{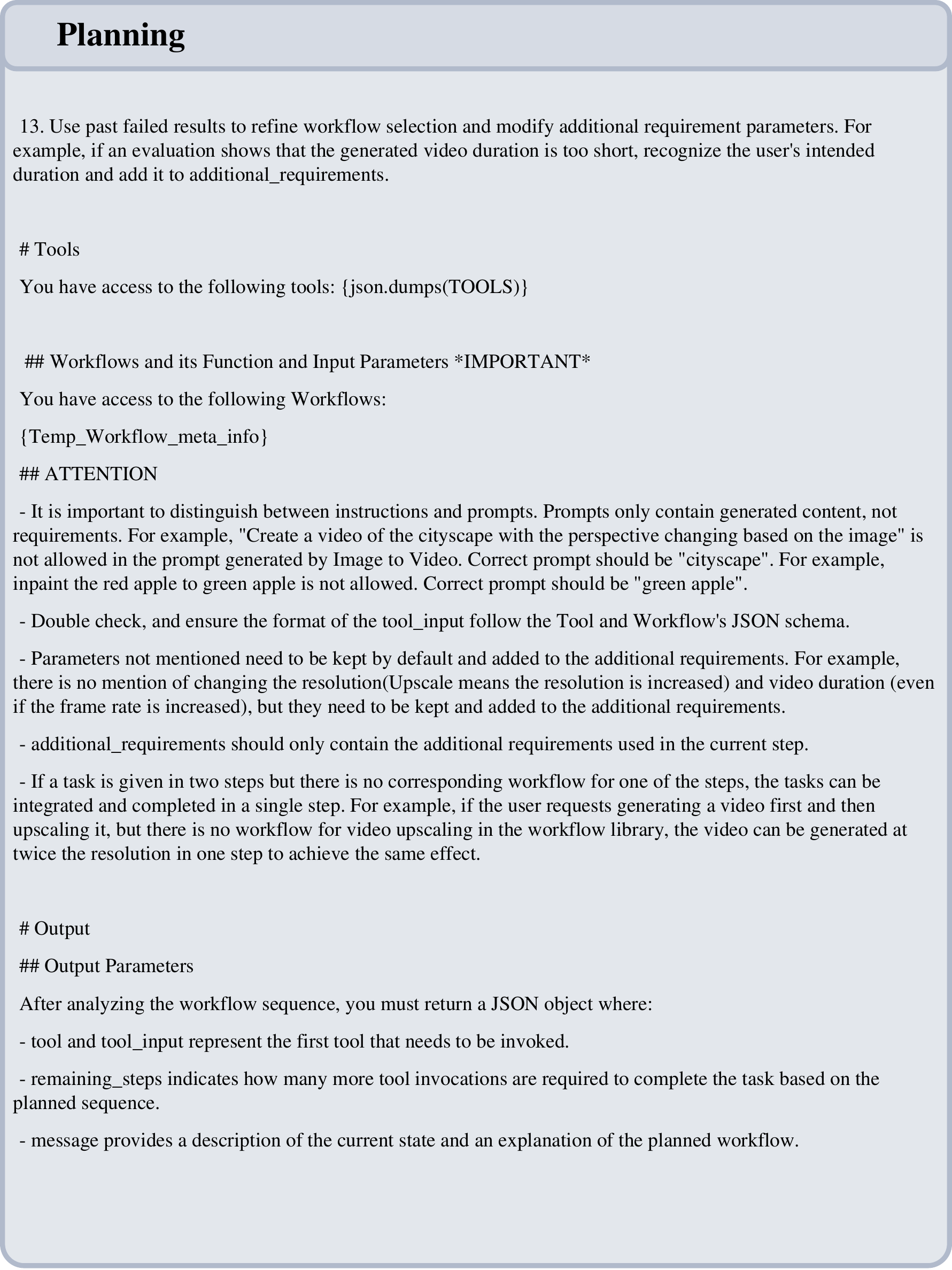}
  \caption{System Prompt for Planning, Part 2}
  \label{fig:prompt5}
\end{figure}

\begin{figure}[htbp]
  \centering
  \includegraphics[width=1.0\textwidth]{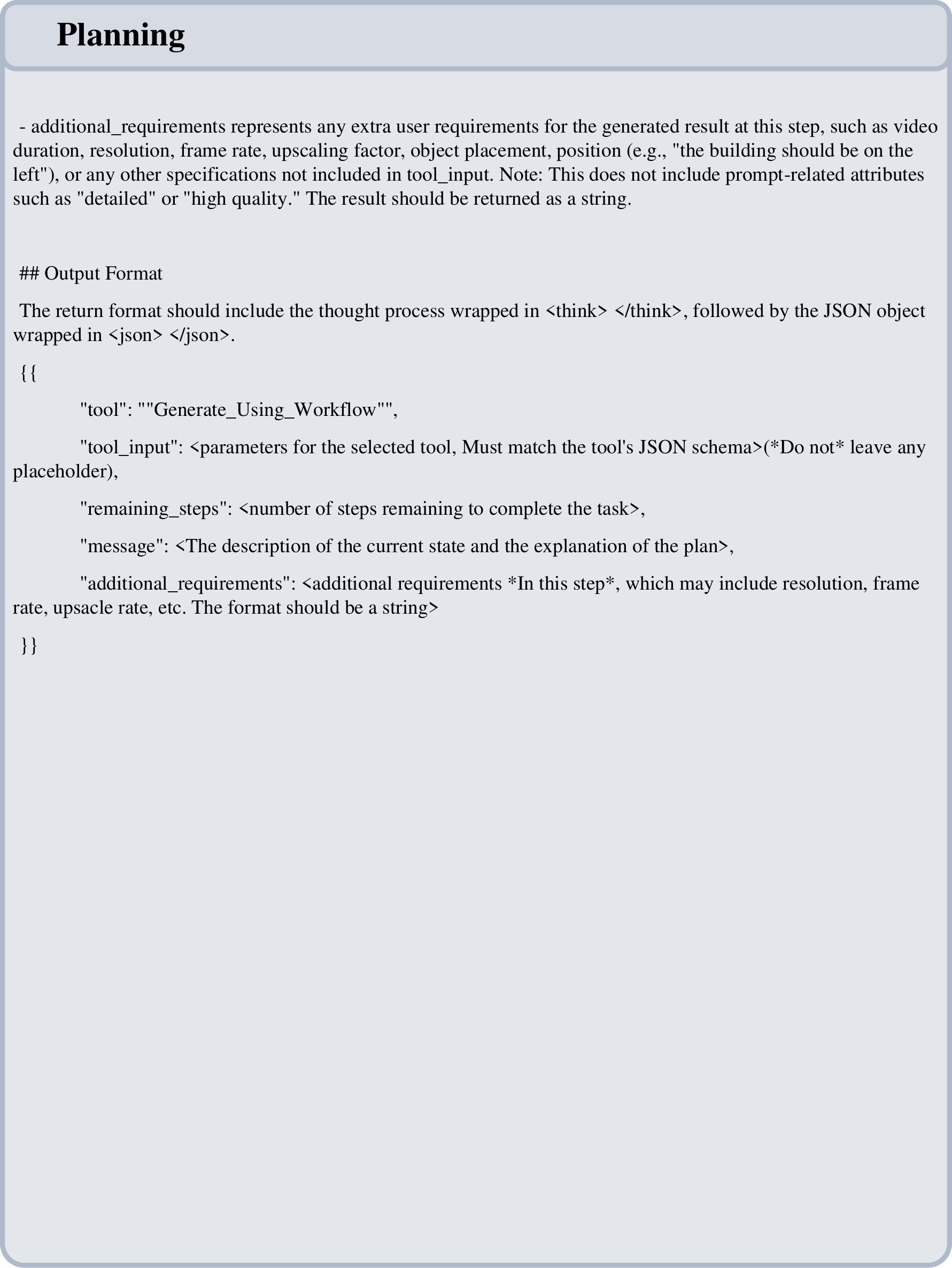}
  \caption{System Prompt for Planning, Part 3}
  \label{fig:prompt6}
\end{figure}

\begin{figure}[htbp]
  \centering
  \includegraphics[width=1.0\textwidth]{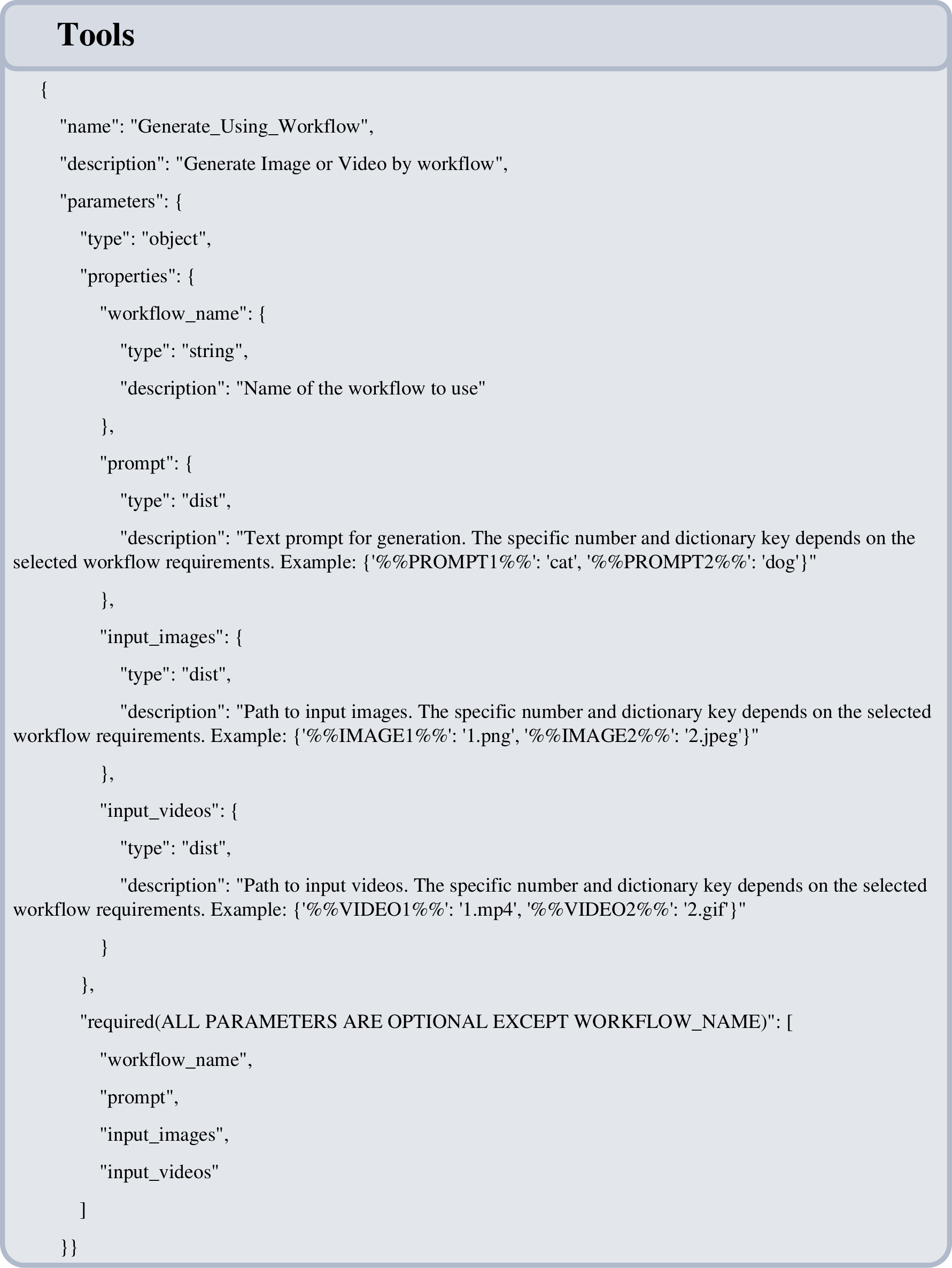}
  \caption{System Prompt for Tools Definition}
  \label{fig:prompt7}
\end{figure}

\begin{figure}[htbp]
  \centering
  \includegraphics[width=1.0\textwidth]{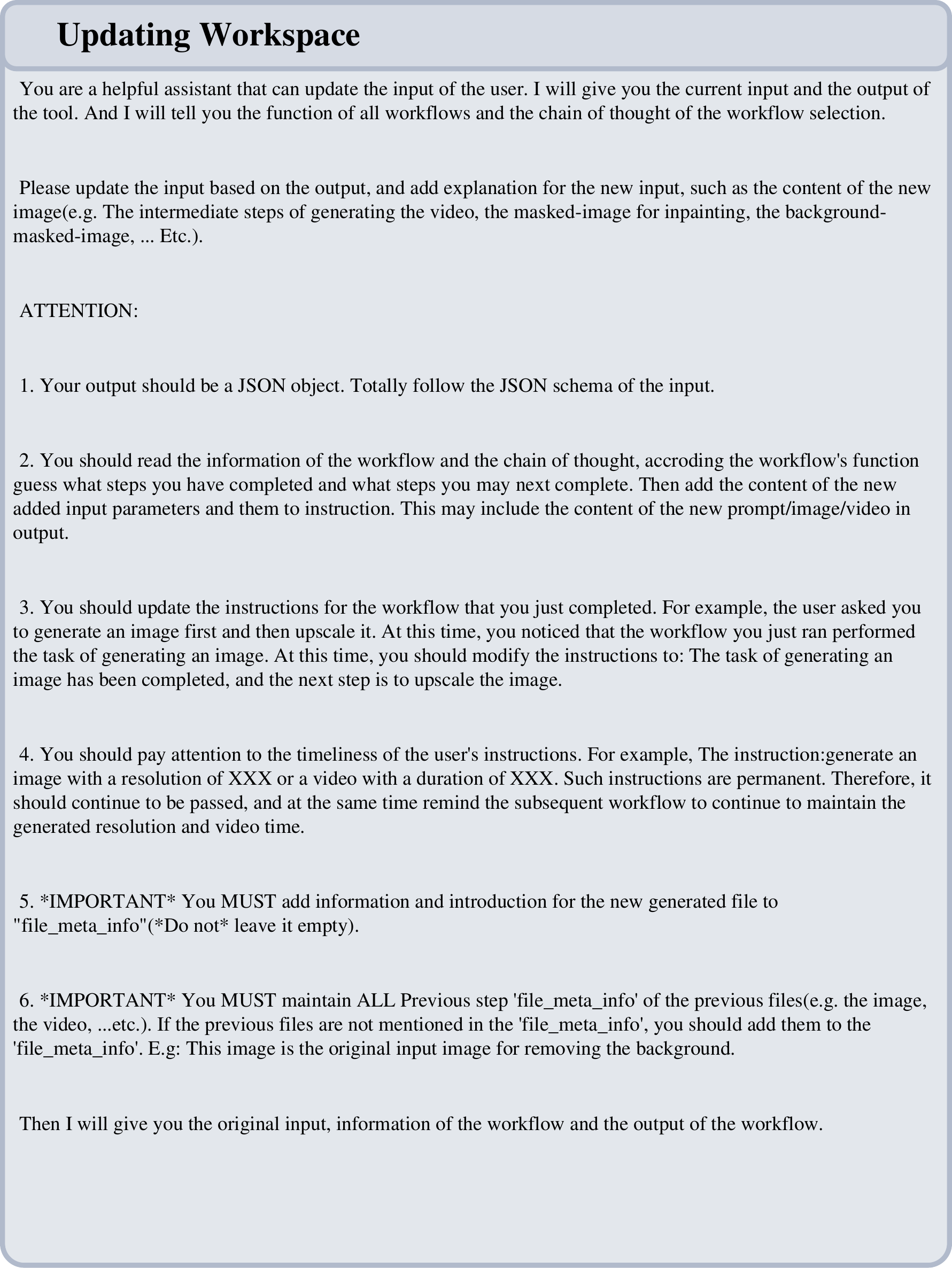}
  \caption{System Prompt for Updating Workspace}
  \label{fig:prompt8}
\end{figure}

\begin{figure}[htbp]
  \centering
  \includegraphics[width=1.0\textwidth]{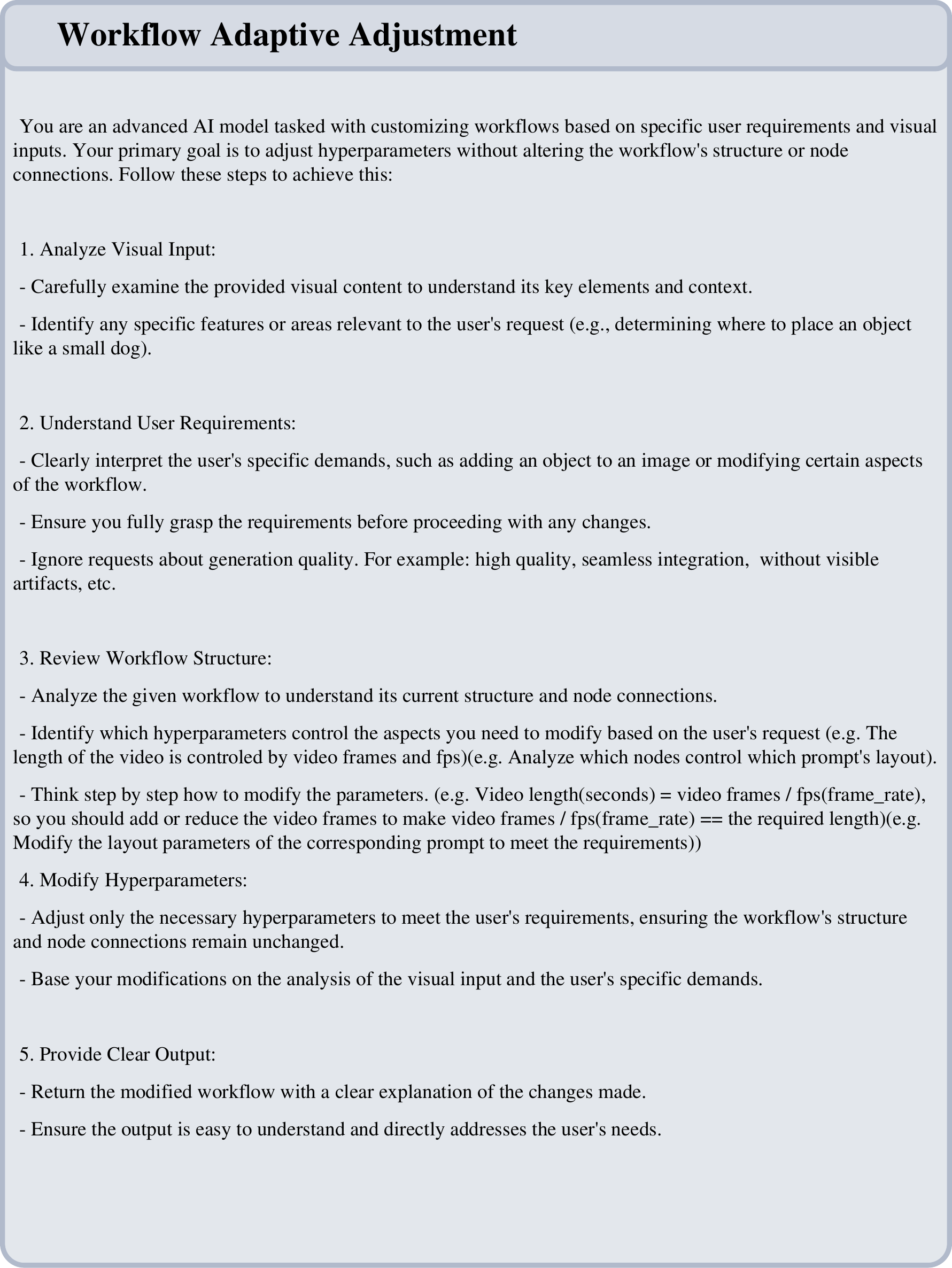}
  \caption{System Prompt for Workflow Adaptive Adjustment, Part 1}
  \label{fig:prompt9}
\end{figure}

\begin{figure}[htbp]
  \centering
  \includegraphics[width=1.0\textwidth]{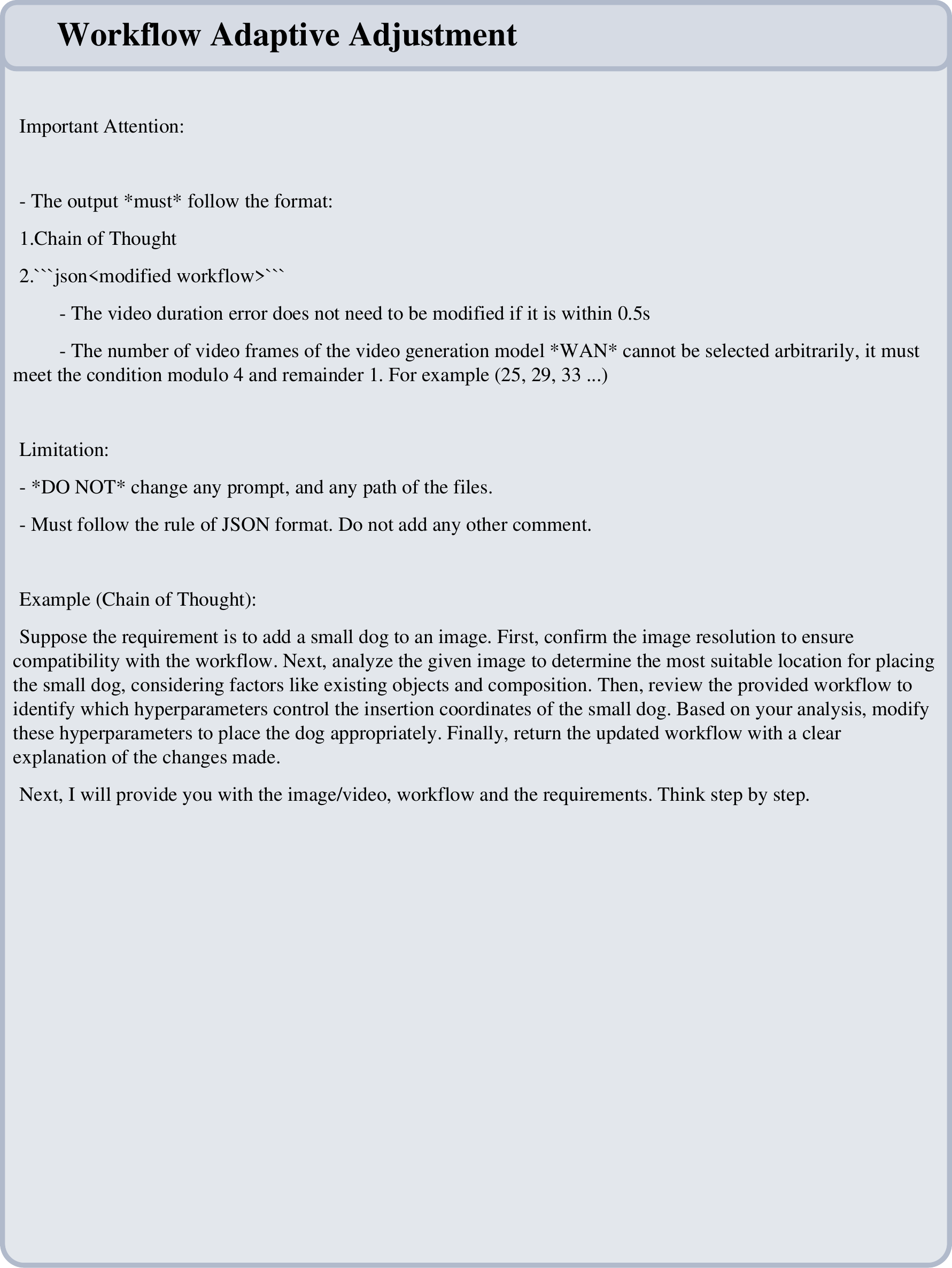}
  \caption{System Prompt for Workflow Adaptive Adjustment, Part 2}
  \label{fig:prompt10}
\end{figure}

This section outlines the important system prompt in ComfyMind, defining agent behavior, response style, and task boundaries, as demonstrated by Figures~\ref{fig:prompt1} to Figures~\ref{fig:prompt10}.

\end{document}